\documentclass[letterpaper]{article}
\usepackage[preprint]{aaai2027}
\usepackage[hyphens]{url}
\usepackage{graphicx}
\usepackage{natbib}
\usepackage{caption}
\usepackage{booktabs}
\usepackage{amsmath}
\usepackage{amssymb}

\title{A Few Neurons Reveal When LLMs Misuse Tools: Sparse Detection \\
and Selective Steering for Reliable Tool Use}

\author{Yutong Ke\equalcontrib,
Ming Yin\equalcontrib,
Chongwen Zhao,
Kaizhu Huang\corresponding}
\affiliations{}

\begin{document}
\maketitle

\begin{abstract}
Agentic LLMs exhibit three consequential tool-use failures: invalid arguments
(\emph{validity}), unnecessary calls (\emph{over-calling}), and omitted calls
when tools are needed (\emph{missing}). We find that a small,
failure-specific set of MLP neurons could distinguish such failure with
linearly separable decision boundaries. Building on this observation, we
introduce \textsc{PRISMS} (\textbf{P}robing \textbf{R}epresentations
\textbf{I}n \textbf{S}upport of \textbf{M}onitoring and
\textbf{S}teering), a closed-loop framework that shares a failure-specific
neuron basis between sparse detection and activation steering.
\textsc{PRISMS} selects contribution-critical MLP neurons and fits an
$L_1$-regularized detector on their activations. Across six models from the
Qwen3, Llama, and Gemma families, over-calling and missing are detected at the
pre-generation prompt boundary with ROC-AUC $0.90$--$1.00$, while validity is
detected from the generated tool-call span with ROC-AUC $0.86$--$0.90$.
These results are achieved with highly sparse readouts: only $1$--$2$ MLP
neurons for missing, $2$--$16$ for over-calling, and approximately $128$ for
validity. These sparse detectors match or outperform dense residual-stream
baselines using $23$--$627\times$ fewer features. The shared neuron basis also
supports bidirectional control over tool-calling behavior, suppressing
unnecessary calls and eliciting omitted ones. \textsc{PRISMS} therefore gates
intervention on predicted failure risk to mitigate the collateral effects of
unconditional steering. Across all six models, \textsc{PRISMS} reduces pooled
over-calling rate by $80\%$ ($0.131 \rightarrow 0.026$) while increasing
tool-required accuracy by $14.2$ percentage points
($0.689 \rightarrow 0.831$). \textsc{PRISMS} thus provides lightweight
failure detection and selective intervention across model families.
\end{abstract}

\section{Introduction}
\label{sec:introduction}

Large language models (LLMs) increasingly act through external tools,
including search engines, code interpreters, databases, and
real-world APIs. Tool use expands what an LLM can know and
execute, but it also introduces failures at the boundary between
reasoning and action. A model may invoke a tool that does not
apply to the request (\emph{over-calling}), fail to invoke an
available tool when needed (\emph{missing}), or select an
applicable tool but generate incorrect argument values
(\emph{validity}). Over-calling wastes latency and API budget,
missing leaves the model to answer beyond its capabilities, and
invalid arguments may trigger unintended actions. 
Reliable agents must therefore improve tool selection and argument
correctness:
they should recognize decision failures before acting and validate
a generated call before external execution~\citep{patil2025bfcl,liu2025toolace,zhang2025xlam}.

\begin{figure}[t]
    \centering
    \includegraphics[width=0.90\columnwidth]{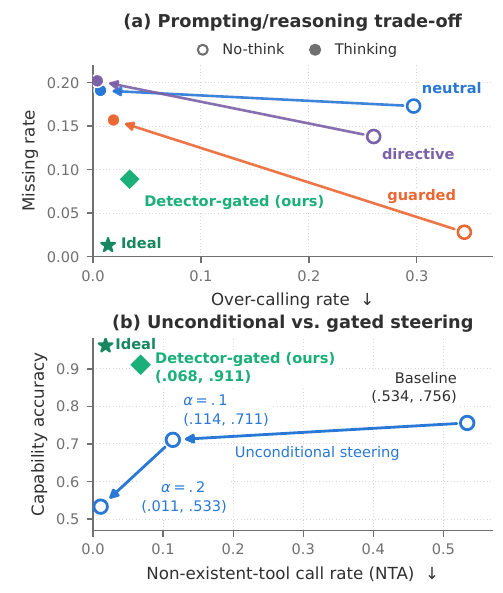}
    \caption{Global controls expose a tool-use trade-off.
    \textbf{(a)} Thinking reduces over-calling but increases missing;
    gated \textsc{PRISMS} (green diamond) approaches the ideal.
    \textbf{(b)} Unconditional steering reduces the
    non-existent-tool call rate (NTA) at a capability cost, whereas
    gated \textsc{PRISMS} improves both.}
    \label{fig:motivation-tradeoffs}
\end{figure}

Recent work has pursued complementary approaches to improving
agentic decision making. Reasoning-centric methods such as ReAct
and Reflexion provide additional deliberation through interleaved
reasoning and action or feedback-induced verbal reflection
\cite{yao2023react,shinn2023reflexion}, but they do not explicitly
monitor whether a particular tool-use failure is present.
\textsc{When2Tool} shows that tool necessity is linearly readable
from hidden states and uses the prediction to encourage tool calling
or abstention \cite{sun2026when2tool}. Yet its all-layer dense
residual readout is high-dimensional and potentially redundant,
while the textual prefix used for control modifies the generation
context. More broadly, stricter prompting, explicit reasoning, and
unconditional activation steering modify the model's overall
tendency to call tools. As
Figure~\ref{fig:motivation-tradeoffs} illustrates, these controls
trade reduced over-calling for increased missing or reduced
legitimate tool use. They therefore shift the global
call-versus-abstain operating point rather than determining which
individual prompts require correction.

Selective correction therefore requires a closed loop with two
separate capabilities: determining \emph{when} a failure is likely
and determining \emph{how} to change the model only then.
Existing work provides pieces of this loop: dense-state monitors
can trigger external execution policies
\citep{healy2026internal}, while activation steering offers an
internal control mechanism
\citep{zou2023representation,turner2023activation,wang2026asa}.
We study the complete loop for three naturally occurring failure
modes---over-calling, missing, and validity---using the models'
own rollouts. We evaluate both failure detection and behavioral
control, distinguishing erroneous calls when no tool is supplied
(NTA) from calls to an inapplicable distractor (DT); their
prompt-weighted combination is pooled over-calling (OC), while
capability accuracy (CA) measures correct calls when a tool is
required.

Related representation-engineering work probes internal
truthfulness and tool-use decisions
\citep{azaria2023lying,wu2026toolcalling,cheng2026knowingdoing}, localizes
hallucination-relevant neurons \citep{gao2025hneuron}, and steers
truthfulness, agent behavior, and tool invocation through sparse
internal directions \citep{li2023iti,sui2026tact,chen2026heading}.
\textsc{PRISMS} extends these ideas to naturally occurring
tool-use failures by coupling compact FFN monitors with
detector-gated correction.

We ask whether these failures can be monitored through compact,
individually addressable FFN features at decision-appropriate
locations, whether strong decodability identifies a causal control
handle, and whether intervention can be applied selectively. Since
probe performance need not imply behavioral use of the encoded
signal \citep{elazar2021amnesic,ilyas2019nonrobust}, we explicitly
separate \emph{reading} from \emph{controlling} a failure. We
address these questions with \textsc{PRISMS}
(\textbf{P}robing \textbf{R}epresentations \textbf{I}n
\textbf{S}upport of \textbf{M}onitoring and \textbf{S}teering),
a novel closed-loop framework that separates \emph{when} to intervene
from \emph{how} to change the model. A sparse detector reads
failure risk from contribution-selected FFN writes, while
independently constructed layerwise directions determine how to
intervene. Across six instruction-tuned models from the Qwen3,
Llama-3.1, and Gemma families, all three failures are linearly
readable when sufficient natural positive examples are available.
On Qwen3, over-calling and missing reach ROC-AUC $0.98$--$1.00$
before generation, while validity reaches $0.86$--$0.90$ from the
generated call span. These results are achieved with highly sparse
readouts: only $1$--$2$ MLP neurons for missing, $2$--$16$ for
over-calling, and approximately $128$ for validity. Under
chain-of-thought reasoning, the most informative decision readout
shifts from the prompt boundary to the reasoning end.

High detectability does not make the selected neurons causally
special: at the main feature budget, matched random subsets
achieve nearly identical detection
($\Delta\mathrm{AUC}\leq 0.011$), and direct intervention on
probe-weighted neurons does not separate from controls.
Contribution-selected layerwise directions, in contrast,
produce a strong bidirectional change in tool-calling and clearly
outperform profile-matched random directions. Closing the loop
then converts this general propensity control into a selective
intervention. Across all six models, detector-gated steering
simultaneously lowers pooled over-calling (OC) and raises capability accuracy (CA), reducing OC by $80\%$ on
average ($0.131\to0.026$) and increasing CA by an average of
$14.2$ percentage points ($0.689\to0.831$).

Our contributions are as follows:
\begin{itemize}
    \item We identify decision-appropriate readouts for three
    naturally occurring tool-use failures: over-calling and missing
    before generation (or at the reasoning end under chain-of-thought),
    and validity over the generated call span. The signals exhibit
    distinct sparsity and depth profiles, with decision failures
    decodable from only a few MLP neurons.

    \item We identify contribution-selected, layerwise steering
    directions that bidirectionally control tool-calling behavior.
    Intervening along these directions suppresses unnecessary calls
    or elicits omitted ones, demonstrating causal control over
    tool-call decisions.

    \item We introduce a detector-gated controller that combines
    sparse monitoring with failure-specific steering. It leaves
    unflagged computation unchanged, suppresses predicted
    over-calling, elicits calls under predicted missing, and
    monitors argument validity before execution.
\end{itemize}

\begin{figure*}[t]
    \centering
    \includegraphics[width=0.85\textwidth]{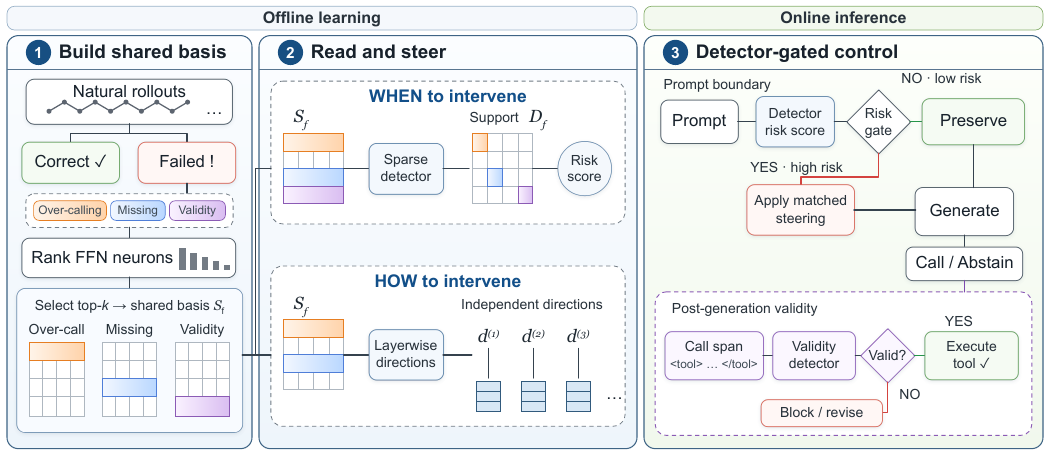}
    \caption{Overview of \textsc{PRISMS}. Natural rollouts define
    a contribution-ranked shared basis $\mathcal{S}_f$. A sparse
    detector determines \emph{when} to intervene, while
    independent layerwise directions determine \emph{how}. The
    risk gate preserves unflagged computation and applies matched
    steering only to predicted failures; validity is checked
    before tool execution.}
    \label{fig:prisms-framework}
\end{figure*}

\section{Method}
\label{sec:method}

\subsection{Overview and Failure-Aware Readouts}
\label{sec:method-overview}

We introduce \textsc{PRISMS}, a closed-loop
framework for monitoring and correcting tool-use failures. We
study three failure modes: \emph{over-calling}, an unnecessary
call when no offered tool applies; \emph{missing}, failure to call
an available, required tool; and \emph{validity}, an applicable
call whose generated arguments disagree with the reference. For
mode $f$, let $\mathcal{H}_f$ and $\mathcal{B}_f$ denote failed
and matched correct rollouts.

Each failure is read where it first becomes observable. Decision
failures are read at the final prompt token without explicit
reasoning and at the token immediately before \texttt{</think>}
when reasoning precedes the action. Validity instead pools the
complete generated call span, including the tool name, argument
keys and values, and delimiters. We write
$\mathcal{T}_f(x,r)$ for the corresponding one-token or call-span
readout in rollout $(x,r)$.

From correct and failed rollouts, \textsc{PRISMS} constructs a
contribution-selected shared basis $\mathcal{S}_f$. An
$L_1$-regularized probe within this basis yields a compact
detector support
$\mathcal{D}_f\subseteq\mathcal{S}_f$, which determines
\emph{when} intervention is needed. Independently constructed
layerwise directions from the full basis determine \emph{how} to
change the model. Over-calling and missing form the online control
loop; validity is monitored after generation and before tool
execution. Figure~\ref{fig:prisms-framework} summarizes this
read--steer pipeline.

\subsection{Contribution-Selected Shared Neuron Basis}
\label{sec:shared-basis}

\paragraph{Neuron writes and contribution.}
Let $h_t^{(\ell)}\in\mathbb{R}^{d_{\mathrm{model}}}$ be the
residual state at token $t$ and layer $\ell$. For neuron
$(\ell,i)$ in a gated FFN, define its scalar activation
$a_{t,i}^{(\ell)}$, down-projection value $v_i^{(\ell)}$, and
residual-stream write $g_{t,i}^{(\ell)}$ by
\begin{equation}
\begin{aligned}
a_{t,i}^{(\ell)}
&:=
\big[
\sigma(h_t^{(\ell)}W_{\mathrm{gate}}^{(\ell)})
\odot(h_t^{(\ell)}W_{\mathrm{up}}^{(\ell)})
\big]_i,
\\
v_i^{(\ell)}
&:=W_{\mathrm{down}}^{(\ell)}[:,i],
\qquad
g_{t,i}^{(\ell)}:=a_{t,i}^{(\ell)}v_i^{(\ell)},
\\
\makebox[0pt][l]{$\displaystyle
\operatorname{MLP}^{(\ell)}(h_t^{(\ell)})
=\sum_i g_{t,i}^{(\ell)}$}
\phantom{c_{t,i}^{(\ell)}}&
\\
c_{t,i}^{(\ell)}
&:=\|g_{t,i}^{(\ell)}\|_2
=|a_{t,i}^{(\ell)}|\|v_i^{(\ell)}\|_2.
\end{aligned}
\label{eq:neuron-contribution}
\end{equation}
This definition follows the FFN key--value-memory view, in which
the output is a weighted sum of value vectors
\citep{geva2021ffn}. The score $c_{t,i}^{(\ell)}$ is also a
column-grouped WANDA analogue: it combines input activation with
down-projection magnitude and equals the norm of the neuron's
actual residual write \citep{sun2024wanda}. Related work uses
WANDA-style attribution to isolate sparse safety-critical regions
\citep{wei2024brittleness}.

\paragraph{Corpus aggregation and global selection.}
For either corpus
$\mathcal{A}\in\{\mathcal{H}_f,\mathcal{B}_f\}$, we average the
contribution over examples and the appropriate readout positions.
With failure and correct selection budgets $k$ and $m$,
respectively,
\begin{equation}
\begin{aligned}
C_{f,i}^{(\ell),\mathcal{A}}
&:=
\mathbb{E}_{(x,r)\sim\mathcal{A}}
\left[
\frac{1}{|\mathcal{T}_f(x,r)|}
\sum_{t\in\mathcal{T}_f(x,r)}
c_{t,i}^{(\ell)}(x,r)
\right],
\\
\mathcal{S}_f
&:=
\operatorname{Top}_{k}
\!\left(C_f^{\mathcal{H}_f}\right)
\cup
\operatorname{Top}_{m}
\!\left(C_f^{\mathcal{B}_f}\right),
\\
\mathcal{S}_f^{(\ell)}
&:=\{i:(\ell,i)\in\mathcal{S}_f\}.
\end{aligned}
\label{eq:shared-basis}
\end{equation}
The ranking is global across all layer--neuron pairs, without a
per-layer quota. The basis can therefore concentrate in layers
that carry more information for failure mode $f$.

\subsection{Sparse Monitoring}
\label{sec:sparse-control}

\paragraph{Signed detector features.}
Basis selection uses unsigned contribution magnitude. Monitoring
instead preserves activation sign and normalizes each write by
the scale of its layer's MLP output:
\begin{equation*}
\begin{aligned}
\phi_{t,i}^{(\ell)}(x,r)
&:=
\frac{
a_{t,i}^{(\ell)}(x,r)\|v_i^{(\ell)}\|_2
}{
\|\operatorname{MLP}^{(\ell)}(h_t^{(\ell)})\|_2+\epsilon
},
\\
z_{f,i}^{(\ell)}(x,r)
&:=
\frac{1}{|\mathcal{T}_f(x,r)|}
\sum_{t\in\mathcal{T}_f(x,r)}
\phi_{t,i}^{(\ell)}(x,r),
\quad(\ell,i)\in\mathcal{S}_f .
\end{aligned}
\label{eq:signed-features}
\end{equation*}
Here $\epsilon>0$ ensures numerical stability, and $z_f(x,r)$
collects these components over $\mathcal{S}_f$. The full
shared-basis probe uses exactly this feature set, not a full-width
residual or MLP representation.

\paragraph{Sparse detector.}
For each failure mode, we fit an $L_1$-regularized logistic probe
within the selected basis:
\begin{equation}
\begin{aligned}
p_f(x,r)
&:=\sigma\!\left(w_f^\top z_f(x,r)+b_f\right),
\\
\mathcal{D}_f
&:=\operatorname{supp}(w_f)
\subseteq\mathcal{S}_f .
\end{aligned}
\label{eq:sparse-detector}
\end{equation}
Only the compact support $\mathcal{D}_f$ is required at inference.

Detector fitting uses training prompts only. The decision
threshold $\tau_f$ and intervention strength $\alpha_f$ are
selected jointly on a prompt-disjoint validation split and frozen
before test evaluation. Among validation settings on the OC--CA
Pareto frontier, we select the largest threshold and then the
smallest strength whose target-metric gain is within $95\%$ of
the best validation gain. Threshold and strength sweeps are
reported in Appendix~\ref{app:calibration-conflicts}.

\subsection{Layerwise Causal Control}
\label{sec:layerwise-control}

Sparse decodability establishes that failure information is
readable, not that probe-supported neurons are causal handles.
We therefore construct control directions from the complete
contribution-selected basis $\mathcal{S}_f$, rather than from
probe weights or $\mathcal{D}_f$ alone, and test their downstream
effect through intervention.

\paragraph{Failure-contrast directions.}
For corpus
$\mathcal{A}\in\{\mathcal{B}_f,\mathcal{H}_f\}$, let
$\bar u_{f,\mathcal{A}}^{(\ell)}$ be the mean residual write of
the selected neurons at layer $\ell$. The normalized
correct-minus-failure direction is
\begin{equation}
\begin{aligned}
\bar u_{f,\mathcal{A}}^{(\ell)}
&:=
\mathbb{E}_{(x,r)\sim\mathcal{A}}
\left[
\frac{1}{|\mathcal{T}_f(x,r)|}
\sum_{t\in\mathcal{T}_f(x,r)}
\sum_{i\in\mathcal{S}_f^{(\ell)}}
g_{t,i}^{(\ell)}(x,r)
\right],
\\
d_f^{(\ell)}
&:=
\bar u_{f,\mathcal{B}_f}^{(\ell)}
-
\bar u_{f,\mathcal{H}_f}^{(\ell)},
\qquad
\widehat d_f^{(\ell)}
:=
\frac{d_f^{(\ell)}}{\|d_f^{(\ell)}\|_2}.
\end{aligned}
\tag*{}
\end{equation}
Each selected layer has its own direction; directions are not
collapsed into one global vector. At an intervention token, the
layer's MLP branch becomes
\begin{equation}
\widetilde{\operatorname{MLP}}^{(\ell)}
\!\left(h_t^{(\ell)}\right)
:=
\operatorname{MLP}^{(\ell)}
\!\left(h_t^{(\ell)}\right)
+
\alpha_f\gamma^{(\ell)}
\widehat d_f^{(\ell)},
\label{eq:layerwise-update}
\end{equation}
where $\alpha_f$ is the validation-selected strength for branch
$f$, and $\gamma^{(\ell)}$ matches the intervention to the scale
of layer $\ell$'s MLP output. The detector therefore answers
\emph{when}; the independently normalized layerwise direction
answers \emph{how}.

\subsection{Detector-Gated Closed Loop}
\label{sec:gated-intervention}

Over-calling and missing use separate detectors, thresholds, and
correcting directions. At the decision readout, the controller
first checks whether
$p_{\mathrm{over}}(x,r)>\tau_{\mathrm{over}}$; if so, it applies
Eq.~\ref{eq:layerwise-update} with the over-calling direction.
Otherwise, if
$p_{\mathrm{miss}}(x,r)>\tau_{\mathrm{miss}}$, it applies the missing
direction. If neither detector fires, the original forward
computation is preserved.

This ordering also resolves the rare case in which both detectors
fire. The two branches target mutually exclusive failures, and a
validation comparison selects over-calling priority as the more
conservative policy. We report conflict frequency and alternative
policies in Appendix~\ref{app:calibration-conflicts}. The
over-calling direction moves the model toward correct abstention;
the missing direction moves it toward correct tool use. They are
independent correct-minus-failure directions, not opposite signs
of one shared vector.

Without explicit reasoning, the gate is read at prompt end; with
reasoning, it is read immediately before \texttt{</think>}. The
validity detector instead reads the generated call span before
execution, after which a downstream agent may execute, block,
flag, or regenerate the call.

\section{Sparse Readouts and Causal Directions}
\label{sec:components}
\label{sec:experimental-setup}

\noindent We first validate the two components behind \textsc{PRISMS}:
whether tool-use failures admit sparse readouts, and whether the
contribution-selected neurons support causal control. The following
protocol is shared across these component analyses and the closed-loop
evaluation in Section~\ref{sec:closed-loop-control}.

\paragraph{Models.}
Our six-model intervention suite spans Qwen3-\{1.7B, 4B, 8B,
14B\}, Llama-3.1-8B-Instruct, and Gemma-3-4B-IT, covering three
model families.
Qwen3's optional thinking mode is evaluated separately in
Section~\ref{sec:think}; all other main results use neutral
prompts without thinking.

\paragraph{Natural-rollout datasets and labels.}
Rather than inject synthetic errors, we label behavior produced by
the models' own stochastic rollouts ($R=8$ for decision failures
and $R=16$ for validity, temperature $0.7$). The \emph{validity}
set uses reference-backed calls from Glaive
\citep{glaive2023function}, ToolACE \citep{liu2025toolace}, and
xLAM \citep{zhang2025xlam}: a parseable, applicable call is
invalid when its argument values disagree with the reference under
AST and normalized-value matching. The \emph{over-calling} set
contains prompts for which no offered tool is applicable; an error
is any emitted call, either with no tool supplied (NTA) or with
only an inapplicable distractor supplied (DT). The \emph{missing}
set contains prompts with an available, required tool; an error is
failure to call it. We construct separate datasets for
over-calling, missing, and validity, each using a prompt-grouped
$70/15/15$ train/validation/test split. Contribution bases and
steering directions are estimated from balanced, equal-sized sets
of failure and correct development prompts. Exact sources,
endpoint thresholds, counts, parsing, and formal label definitions
appear in Appendix~\ref{app:experimental-setup}.

\paragraph{Metrics.}
Monitoring is evaluated with held-out ROC-AUC and average
precision (AP). Behavioral control reports NTA and DT error rates
separately, capability accuracy (CA) on tool-required prompts, and
their prompt-count-weighted pooled over-call rate
\begin{equation}
\mathrm{OC}=
\frac{n_{\mathrm{NTA}}r_{\mathrm{NTA}}+
      n_{\mathrm{DT}}r_{\mathrm{DT}}}
     {n_{\mathrm{NTA}}+n_{\mathrm{DT}}}.
\label{eq:oc}
\end{equation}
Thus OC measures erroneous calling when no valid tool is available,
whereas CA measures preservation or recovery of legitimate tool
use. Direct-steering and supplementary tables retain NTA and DT
separately even when the main gate figure reports OC.

\paragraph{Baselines and controls.}
We reproduce the closest dense hidden-state detectors on matched
data and models: the final-layer three-position MLP of
\citet{healy2026internal} for validity and \textsc{When2Tool}'s
last-input-token linear probe \citep{sun2026when2tool} for tool
necessity. For the contribution basis, we use
$k=m=\lceil0.03N\rceil$ over the $N$ layer--neuron pairs; this
$3\%$ budget is an experimental setting, not part of the method.
Controls include prompt-only and Reason-then-Act baselines,
label-shuffled and layer-profile-matched random directions,
probe-selected neurons, shuffled gates, random vectors routed by
the real gate, and degeneration screening. Complete
implementations, generation parameters, operating-point
selection, and screening criteria are deferred to
Appendix~\ref{app:experimental-setup}.

\paragraph{Validation-only operating points.}
All detector thresholds and steering strengths are selected on
prompt-disjoint validation data and then frozen for the test set.
We search their joint OC--CA trade-off and conservatively choose
the largest threshold and smallest strength within $95\%$ of the
best validation gain for the target branch. The validation sweeps
give a broad stable region around
$\alpha_{\rm over}=\alpha_{\rm miss}=.2$ across all six models;
neighboring-parameter sensitivity and dual-detector conflict
audits appear in Appendix~\ref{app:calibration-conflicts}.

\paragraph{One basis for everything.} All experiments share a single neuron basis per model
and signal. Each MLP neuron is scored by its \emph{contribution}
$c^i_l=\overline{|a^i_l|}\cdot\|W_{\text{down}}[:,i]\|$ on an error corpus $H$ and a correct
corpus $B$ drawn from the model's own rollouts; in the main setting, the basis is the
\emph{union} obtained with $k=m=\lceil0.03N\rceil$. Detection fits an L1-logistic probe on \emph{signed}
per-token contribution features ($z\cdot\|W_{\text{down}}\|/\|\text{mlp\_out}\|$) restricted
to this set; steering perturbs along $d=p^{B}-p^{H}$ built on the \emph{same} set; the gated
repair combines the two. Selection corpora are drawn from training prompts only, and all
test prompts are disjoint from them. The six-model intervention
suite comprises Qwen3-\{1.7B, 4B, 8B, 14B\},
Llama-3.1-8B-Instruct, and Gemma-3-4B-IT.
Unless stated otherwise, results use the neutral prompt and no-think setting.

\subsection{Sparse Failure Readouts}
\label{sec:detect}
\begin{table}[t]
\centering
\footnotesize
\setlength{\tabcolsep}{3pt}
\caption{Detection on the unified basis (held-out test; neutral/no-think).
Cells report ROC-AUC (AP). Decision signals are read pre-generation per
prompt; validity is read over the generated call span per rollout.}
\label{tab:detection}
\begin{tabular}{@{}lccc@{}}
\toprule
Model & over-calling & missing & validity \\
\midrule
Qwen3-1.7B & 0.980 (0.92) & 0.996 (0.98) & 0.889 (0.76) \\
Qwen3-4B   & 0.994 (0.99) & 1.000 (1.00) & 0.886 (0.75) \\
Qwen3-8B   & 0.995 (0.95) & 1.000 (1.00) & 0.895 (0.77) \\
Qwen3-14B  & 0.994 (0.99) & 0.997 (0.99) & 0.856 (0.71) \\
Llama-3.1-8B-Inst.
  & 0.996 (0.98) & 0.895 (0.79) & 0.856 (0.72) \\
Gemma-3-4B
  & 0.908 (0.84) & 0.927 (1.00) & 0.868 (0.75) \\
\bottomrule
\end{tabular}
\end{table}

Table~\ref{tab:detection} reports held-out detection at one operating point for all
models. On Qwen3, over-calling and missing reach ROC-AUC $0.98$--$1.00$, with
over-calling AP $0.92$--$0.99$. Llama reaches AUC/AP $0.996/0.981$ for
over-calling and $0.895/0.79$ for missing; Gemma reaches $0.908/0.84$ and
$0.927/1.00$, respectively. Validity is moderately decodable across all three
families, with AUC $0.856$--$0.895$. Thus, decision failures are strongly readable
before generation, while validity remains readable from the generated call span.

We deliberately make \emph{no} claim here that the contribution-selected basis detects
better than a size- and profile-matched random one: at a $3\%$ budget the readable signal is
redundant enough that detection cannot separate selections ($\Delta$AUC vs.\ random
$\le+0.011$). Whether the selected set is \emph{special} is a causal question, answered by
the steering and gating experiments (Sections~\ref{sec:steer} and~\ref{sec:gate}).

Signed activations are important: magnitude-only features reduce validity AUC by
$0.02$--$0.04$ (8B: $0.854\to0.891$ with sign restored), showing
that part of the validity signal is direction-encoded.

\subsection{Sparsity, Independence, \& Readout Location}
\label{sec:sparse}
\paragraph{Sparsity.} Table~\ref{tab:sparsity} tracks detection as the budget shrinks from
the full selected basis ($5{,}587$--$30{,}556$ features, or $3.1$--$4.6\%$ of the
all-layer MLP feature space) to a per-layer-equivalent budget of one. At $K{=}1$
($35$--$59$ actual union features), over-calling and missing retain near-full AUC
in almost every cell, whereas validity degrades more visibly. A modest increase in
$K$ largely recovers validity. An unconstrained L1 search over raw activations puts
the signal-level limit lower still
($1$--$2$ neurons for missing, $2$--$16$ for over-calling, $\sim$$128$ for validity),
so sparsity is a property of the \emph{signal}; the basis buys causal addressability at the
price of a few dozen features. Consistently with the framing above, the sparse end shows no
detection advantage over matched random subsets---specialness is established causally, not
by AUC deltas. Complete six-model results appear in
Appendix~\ref{app:sparsity-results}.

\begin{table}[t]
\centering
\footnotesize
\setlength{\tabcolsep}{2.2pt}
\caption{Detection AUC as the unified-basis budget shrinks for
representative Qwen, Llama, and Gemma models. Full-basis cells also
give the selected-feature count; other cells give AUC (union count).
The complete six-model table is in Appendix~\ref{app:sparsity-results}.}
\label{tab:sparsity}
\begin{tabular}{@{}llccc@{}}
\toprule
signal & budget & Qwen-4 & Llama-8 & Gemma-4 \\
\midrule
overcall & full basis
 & \shortstack{.994\\13,440} & \shortstack{.996\\19,084}
 & \shortstack{.908\\13,205} \\
 & $K{=}1$ & .996 (45) & .981 (52) & .966 (40) \\
\midrule
missing & full basis
 & \shortstack{1.00\\15,736} & \shortstack{.895\\16,587}
 & \shortstack{.927\\12,262} \\
 & $K{=}1$ & 1.00 (55) & .972 (43) & .811 (37) \\
\midrule
validity & full basis
 & \shortstack{.886\\11,126} & \shortstack{.856\\14,251}
 & \shortstack{.868\\11,184} \\
 & $K{=}4$ & .841 (152) & .826 (136) & .816 (146) \\
 & $K{=}1$ & .780 (41) & .809 (35) & .765 (34) \\
\bottomrule
\end{tabular}
\end{table}

\paragraph{Restricted overlap diagnostic.} In the fixed six-layer
Qwen3-4B subspace used for the original diagnostic, the three
failure-specific top-$K$ sets show no excess overlap beyond chance
(Table~\ref{tab:overlap}). Pairwise overlap is zero for $K\le200$, below
expectation at $K{=}500$, and near chance at $K{=}1{,}000$ ($18$ observed
vs.\ $17.1$ expected); the three-way intersection is empty in this
restricted analysis. The complete all-layer audit finds small,
configuration-sensitive shared components while preserving very low
Jaccard overlap (Appendix Tables~\ref{tab:overlap-all-layer} and
\ref{tab:overlap-ranking-sensitivity}).

\begin{table}[t]
\centering
\small
\caption{Restricted six-layer three-signal overlap vs.\ chance
(Qwen3-4B, common $58{,}368$-feature space). Within this diagnostic,
overlap never exceeds the hypergeometric expectation and the
three-way intersection is empty.}
\label{tab:overlap}
\begin{tabular}{llccc}
\toprule
$K$ & pair & obs. & exp. & obs/exp \\
\midrule
100$\sim$200 & all three pairs & 0 & $\le$0.69 & 0.00 \\
500  & validity\,$\times$\,over-call & 1 & 4.28 & 0.23 \\
500  & other two pairs & 1 & 2.49 & 0.40 \\
1,000 & validity\,$\times$\,over-call & 18 & 17.13 & 1.05 \\
1,000 & validity\,$\times$\,missing & 4 & 4.99 & 0.80 \\
1,000 & over-call\,$\times$\,missing & 2 & 4.99 & 0.40 \\
\bottomrule
\end{tabular}
\end{table}

\begin{figure}[t]
\centering
\includegraphics[width=0.8\columnwidth]{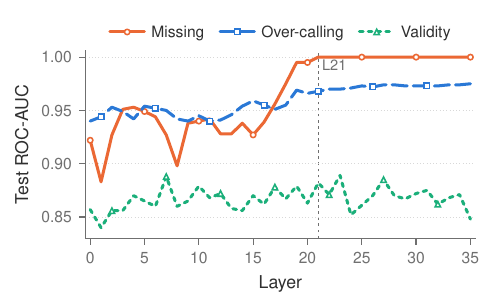}
\caption{Per-layer detection AUC, all $36$ layers (Qwen3-4B). Two regimes: the decision
signals are late-localized---missing snaps to $1.000$ at L21 and stays there; over-calling
climbs from $\approx$$0.94$ to $\approx$$0.97$ after L19---while validity is layer-flat
($0.85$--$0.89$ at every depth, no dominant layer).}
\label{fig:perlayer}
\end{figure}

\paragraph{Depth.} The sets also sit at different depths (Figure~\ref{fig:perlayer}). The
decision signals are \emph{late-localized}---missing's top-$100$ neurons lie entirely in the
late half (mean relative depth $0.87$), its per-layer AUC snapping to $1.000$ at L21;
over-calling is intermediate ($69\%$ late). Validity is \emph{layer-flat} ($43\%$ late, no
dominant layer). Quantified as early$\to$late-half \emph{error} reduction (AUC ranges are
distorted by ceiling effects): $97.0\%$ missing, $44.2\%$ over-calling, $4.4\%$ validity.
The steering direction concentrates in the same place: layers $21$--$27$ carry $\sim$$74\%$
of $\|d\|$. The call/no-call decision crystallizes near the output; judging an emitted
call's correctness does not.

\subsection{Contribution-Selected Neurons Enable Causal Control}
\label{sec:steer}
Detection could not distinguish the contribution basis from random; intervention can.
Steering the \emph{probe's} own top-weight neurons never separates from controls
(permutation $p=0.238$), which motivated selecting by contribution in the first place.
Along the basis direction $d$, additive steering moves over-calling monotonically and
bidirectionally ($+d$ suppresses tool-calling, $-d$ elicits it; e.g.\ 4B non-existent-tool
hallucination $0.534\to0.000$ and $\to0.886$ at the dose extremes), while \emph{validity
does not move in either direction}---mirroring its diffuse, sign-encoded readout.

The decisive control compares the real direction with a
\emph{matched-random} one that preserves neuron count and
per-layer profile but randomizes neuron identity
(Table~\ref{tab:steer}). Across Qwen, the real direction removes
$79$--$100\%$ of over-calling while the matched-random direction
is largely ineffective; on 4B, random steering is even worse than
baseline ($0.580$ vs.\ $0.534$). Thus causal specificity lies in
the selected neurons, not merely their number or depth profile.
The effect extends beyond Qwen: on Llama, the real direction
reduces DT from $.228$ to $.101$, compared with $.201$ for the
matched control; on Gemma, it reduces DT from $.054$ to zero,
whereas the control raises it to $.144$. Unconditional steering
can nevertheless reduce capability at stronger doses, motivating
the detector gate below. Post-hoc screening found no repetition
or overlength degeneration.

\begin{table}[t]
\centering
\footnotesize
\setlength{\tabcolsep}{2.5pt}
\caption{Real vs.\ matched-random steering direction (suppress arm, relative $\alpha$).
\textbf{NTA} is erroneous calling when no tool is supplied; \textbf{DT} is erroneous
calling when only an inapplicable distractor tool is supplied; \textbf{CA} is capability
accuracy on tool-required prompts. Random preserves neuron count and layer profile.}
\label{tab:steer}
\begin{tabular}{@{}lccc@{}}
\toprule
 & base & real $d$ & random $d$ \\
model ($\alpha$) & \multicolumn{3}{c}{NTA$\downarrow$\,/\,DT$\downarrow$\,/\,CA$\uparrow$} \\
\midrule
Qwen3-1.7B (.1) & .114/.079/.867 & .011/.034/.600 & .080/.067/.867 \\
Qwen3-4B (.1)   & .534/.011/.756 & .114/.000/.711 & .580/.022/.756 \\
Qwen3-8B (.1)   & .091/.011/.956 & .011/.011/.911 & .080/.022/.933 \\
Qwen3-14B (.2)  & .420/.011/.767 & \textbf{.023/.000}/.581 & .239/.000/.698 \\
Llama-8B (.1)   & .010/.228/.685 & .009/.101/\textbf{.664} & .016/.201/.589 \\
Gemma-3-4B (.2) & .015/.054/.078 & \textbf{.005/.000}/.000 & .020/.144/.255 \\
\bottomrule
\end{tabular}
\end{table}

\section{Closed-Loop Tool-Use Control}
\label{sec:closed-loop-control}
Having established sparse failure readouts and contribution-specific causal
directions separately, we now evaluate their combination as a detector-gated
controller and test its robustness across models and reasoning modes.

\subsection{Main Detector-Gated Steering Results}
\label{sec:gate}
\begin{figure}[t]
\centering
\includegraphics[width=0.86\columnwidth]{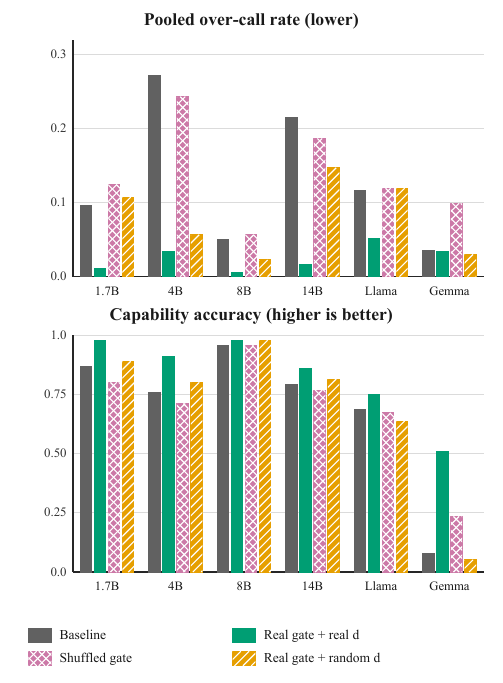}
\caption{Bidirectional detector-gated steering. Pooled over-call rate
(OC$\downarrow$) combines NTA and DT by prompt count; CA$\uparrow$
measures correct calls on tool-required prompts.
The full NTA/DT decomposition appears
in Appendix Table~\ref{tab:gate-full}.}
\label{fig:gate-strength}
\end{figure}

Unconditional steering is a \emph{propensity} knob---suppressing hallucinated calls also
suppresses legitimate ones. Because the detector reads the same basis the vector writes, the
fix is architectural: a bidirectional gate that intervenes only where the probe predicts a
failure ($p_{\text{over}}>\tau_{\text{over}} \Rightarrow
+\alpha_{\text{over}}d_{\text{over}}$ to suppress;
$p_{\text{miss}}>\tau_{\text{miss}} \Rightarrow
+\alpha_{\text{miss}}d_{\text{miss}}$ to elicit), leaving every other forward pass
untouched.

Figure~\ref{fig:gate-strength} reports the pooled OC metric from
Equation~\ref{eq:oc}; NTA and DT remain separate in the appendix.
Operating points are selected on validation data and frozen for
test, with $\alpha=.2$ lying in a broad stable region
(Appendix~\ref{app:calibration-conflicts}). Across all six models,
the complete controller lowers macro OC by $80\%$
($.131\to.026$) and raises CA by $14.2$ points
($.689\to.831$). It improves different regimes: Qwen3-4B moves
from $.271/.756$ to $.034/.911$ OC/CA; Llama-8B reduces DT from
$.228$ to $.096$ while raising CA from $.685$ to $.751$; and
Gemma-3-4B keeps its low OC nearly fixed while recovering CA from
$.078$ to $.508$. Thus gating can suppress erroneous calls,
recover required calls, or do both.

Controls show that both routing and direction matter. On
Qwen3-4B, shuffled gating reaches $.243/.711$ OC/CA and a random
direction $.056/.800$, versus $.034/.911$ for the complete
controller. Neither control matches the real controller on Llama
or Gemma either. These failures rule out intervention frequency
and indiscriminate tool-call elicitation; complete subtype results
and prompt counts appear in Table~\ref{tab:gate-full}.

\subsection{Comparison with Prior Tool-Use Methods}
\label{sec:compare}
Table~\ref{tab:baselinecompare} compares our sparse probe with
faithfully reproduced detectors in three matched Qwen3-4B
comparisons. On our over-calling corpus, \textsc{PRISMS} reaches
$0.994$ AUC versus $0.978$ for \textsc{When2Tool}'s final-input
dense probe. On \textsc{When2Tool}'s original tool-necessity
benchmark and prompt wrapper, \textsc{PRISMS} reaches $0.934$
versus $0.880$ for its all-layer dense probe. On our
natural-rollout validity data, \textsc{PRISMS} reaches $0.886$,
versus $0.869$ for \textsc{When2Tool} and $0.856$ for the
three-position MLP of \citet{healy2026internal}. Across these settings,
\textsc{PRISMS} uses $23$--$627\times$ fewer individually
addressable inputs. Substituting our sparse gate into
\textsc{When2Tool}'s unchanged prefill controller raises robust
accuracy by $1.7$--$2.2$ points at matched call rates. A
natural-over-calling suppress direction also transfers without
refitting, reducing unnecessary calls by $18.5$ points against
matched random at $\alpha=.4$ with no degeneration. Full
frontiers, controls, and per-source results appear in
Appendix~\ref{app:baseline-comparison}.

\begin{table}[t]
\centering
\footnotesize
\setlength{\tabcolsep}{3pt}
\caption{Matched Qwen3-4B detector comparisons. Input dim.\ is the
number of features presented to each probe.}
\label{tab:baselinecompare}
\begin{tabular}{@{}llrc@{}}
\toprule
Signal & Method & Input dim. & AUC \\
\midrule
Over-calling
& \textsc{When2Tool}
& 2,560 & 0.978 \\
& \textsc{PRISMS} (ours)
& $\sim$64 & \textbf{0.994} \\
\midrule
Necessity
& \textsc{When2Tool}
& 94,720 & 0.880 \\
& \textsc{PRISMS} (ours)
& 151 & \textbf{0.934} \\
\midrule
Validity
& \textsc{When2Tool}
& 92,160 & 0.869 \\
& Healy et al.
& 7,680 & 0.856 \\
& \textsc{PRISMS} (ours)
& 330 & \textbf{0.886} \\
\bottomrule
\end{tabular}
\end{table}

\subsection{Reasoning-Mode Monitoring}
\label{sec:think}
Reasoning changes where a decision should be read. We rebuild the
unified basis at the token before \texttt{</think>} and compare it
with the prompt-end readout; the response-side tag is present on
$\approx100\%$ of rollouts, avoiding contamination from an empty
prompt-side \texttt{<think></think>}. Across every Qwen3 setting,
the readable decision shifts toward the reasoning end
(Appendix Table~\ref{tab:think-full}). The migration is
signal-dependent. On Qwen3-4B, over-calling AP rises from $0.083$
at prompt end to $0.676$ at think end ($\Delta$AP $+0.59$), with
large gaps at every scale: the decision is formed during
reasoning. Missing migrates more softly ($\Delta$AP $+0.07$ to
$+0.18$), because prompt-end AP is already $0.81$--$0.93$.
Thus a reasoning-mode monitor should read over-calling at
\texttt{</think>}, while missing is partly visible earlier.

Thinking also sharpens validity without relocating it: using the
same generated-call span, row-AUC increases from $0.889$ to
$0.913$ (1.7B) and from $0.886$ to $0.934$ (4B). A back-off sweep
provides a temporal view of the over-calling signal. On Qwen3-1.7B
it is detectable at least $100$ tokens before \texttt{</think>}
but remains diffuse; moving backward from the commitment point,
AP falls from $1.00$ to $0.90$ while detector support expands from
$38$ to $714$ neurons. The signal therefore crystallizes as the
model commits, consistent with thinking's observed abstention
bias.

\begin{table}[t]
\centering
\footnotesize
\setlength{\tabcolsep}{3.5pt}
\caption{Representative Qwen3-4B reasoning results. Decision cells
compare prompt-end with think-end AUC/AP; validity compares no-think
with thinking call-span AUC. Complete results are in Appendix
Table~\ref{tab:think-full}.}
\label{tab:think}
\begin{tabular}{@{}lccc@{}}
\toprule
Signal & earlier & later & $\Delta$ \\
\midrule
Over-calling & .609 / .083 & \textbf{.789 / .676} & $\mathbf{+.59}$ AP \\
Missing      & .970 / .831 & \textbf{.991 / .948} & $+.12$ AP \\
Validity     & .886        & \textbf{.934}        & $+.048$ AUC \\
\bottomrule
\end{tabular}
\end{table}

\subsection{Additional Experiments}
\label{sec:additional-experiments}
Complete results and implementation details are provided in the
supplement:
\begin{itemize}
    \setlength{\topsep}{2pt}
    \setlength{\itemsep}{0pt}
    \setlength{\parsep}{0pt}
    \setlength{\parskip}{0pt}
    \item \textbf{Complete experimental setup and results
    (Appendix~\ref{app:experimental-setup}):} Models, data,
    controls, reasoning, calibration, sparsity, and per-model
    intervention results.
    \item \textbf{Reproducibility details
    (Appendix~\ref{app:reproducibility-details}):} Hyperparameter
    ranges, seeds, compute, runs, uncertainty, and statistical tests.
    \item \textbf{Prior-method comparisons and transfer
    (Appendix~\ref{app:baseline-comparison}):} Matched detectors,
    gate substitution, and cross-corpus steering.
    \item \textbf{Extended related work
    (Appendix~\ref{app:related-work}):} Tool-use evaluation,
    probing, sparse neurons, and activation steering.
\end{itemize}

\section{Conclusion}

We introduced \textsc{PRISMS}, a closed-loop framework for
monitoring and selectively correcting naturally occurring
tool-use failures. Across model families, compact FFN feature
sets reliably detect over-calling and missing before generation
and validity errors from the generated call span. Sparse
decodability alone does not identify causal neurons; effective
control instead requires contribution-selected layerwise
directions. Gating these directions with failure-specific
detectors suppresses predicted over-calling and elicits omitted
calls while leaving unflagged computation unchanged, whereas
validity remains a post-generation safety check before tool
execution. These results demonstrate that sparse internal
monitoring can provide an effective trigger for selective,
failure-aware tool-use correction.

\clearpage
\bibliography{refs}

\clearpage
\appendix
\setcounter{table}{0}
\setcounter{figure}{0}
\setcounter{equation}{0}
\renewcommand{\thetable}{S\arabic{table}}
\renewcommand{\thefigure}{S\arabic{figure}}
\renewcommand{\theequation}{S\arabic{equation}}
\section{Complete Experimental Setup}
\label{app:experimental-setup}

\subsection{Evaluated Models and Sampling}
We evaluate six models spanning Qwen3-1.7B, Qwen3-4B, Qwen3-8B,
Qwen3-14B, Llama-3.1-8B-Instruct, and Gemma-3-4B-IT. Qwen3 exposes an optional
\emph{thinking} mode; the other families are no-think only. We
sample $R=8$ generations for the decision signals and $R=16$ for
validity at temperature $0.7$, using each model's own chat
template. All final results use held-out test prompts, with
prompt-level disjointness enforced across training, validation,
and test splits.

\subsection{Baseline Implementations}

\paragraph{Validity: internal-representation tool-hallucination
detection \citep{healy2026internal}.}
Healy et al.\ detect tool-call hallucinations using a two-layer
MLP over final-layer residual states extracted at three positions:
the function token, the mean argument-span representation, and
the end of the call,
\[
h[t_{\mathrm{func}}]
\,\|\,\overline{h[T_{\mathrm{args}}]}
\,\|\,h[t_{\mathrm{end}}].
\]
Their labels are generated through mask-and-replace corruption of
ground-truth calls, whereas ours are derived from errors naturally
produced by model rollouts.

\paragraph{Over-calling: pre-generation tool-necessity probing
and steering \citep{sun2026when2tool}.}
Sun et al.\ read the last-input-token residual state with a linear
probe to predict whether a tool is needed. Although their label
captures ground-truth tool necessity rather than naturally
observed over-calling, it provides the closest pre-generation
baseline for our decision signal. Their \textsc{Probe\&Prefill}
method controls tool use by inserting a steering sentence into the
response instead of intervening on internal activations. We
reproduce their dense residual-state probe on our over-calling
data and additionally compare the two feature spaces on the
\textsc{When2Tool} benchmark.

\subsection{Dataset Construction and Processing}

\textbf{Validity} data derive from four single-turn
function-calling sources with ground-truth calls: Glaive
\citep{glaive2023function}, ToolACE \citep{liu2025toolace}, and
xLAM \citep{zhang2025xlam}; BFCL is collected but excluded from
pooled validity scoring because its near-zero error rate ($3.5\%$)
would allow a probe to rank source signature instead of call
validity. \textbf{Decision} data are built from tool-necessity
scenarios. Over-calling prompts have no suitable available tool
and comprise \texttt{non\_existent\_tool} (no tools offered) and
\texttt{non\_existent\_tool\_with\_distractor} (one plausible but
wrong tool). Missing examples are capability prompts for which a
suitable tool exists and should be called.

\paragraph{Formal natural-rollout labels.}
For each prompt $x$, we sample responses
$\{\hat{y}_x^{(r)}\}_{r=1}^{R}$ and parse an emitted tool call
$c_x^{(r)}$ when present. Validity is evaluated only for
parseable calls against the reference call $c_x^\star$:
\begin{equation}
y_{\mathrm{val}}^{(r)}
:=
\mathbf{1}\!\left[
\operatorname{args}(c_x^{(r)})
\not\equiv
\operatorname{args}(c_x^\star)
\right].
\end{equation}
For decision failures, the natural label combines tool
applicability with the model's emitted action:
\begin{equation}
y_{\mathrm{over}}^{(r)}
:=
\mathbf{1}\!\left[
c_x^{(r)}\neq\varnothing
\land \operatorname{applicable}(x)=0
\right],
\end{equation}
\begin{equation}
y_{\mathrm{miss}}^{(r)}
:=
\mathbf{1}\!\left[
c_x^{(r)}=\varnothing
\land \operatorname{applicable}(x)=1
\right].
\end{equation}
These labels preserve errors generated by the model itself rather
than corrupting a correct reference call.

\paragraph{Stable-endpoint decision labels.}
The decision readout is taken before generation and is therefore
shared by all stochastic rollouts of the same prompt. We avoid
assigning conflicting labels to this shared representation by
summarizing the $R=8$ rollout labels at the prompt level. For
$f\in\{\mathrm{over},\mathrm{miss}\}$, let
\[
s_f(x):=\sum_{r=1}^{R}y_f^{(r)}(x).
\]
We retain only the two stable endpoints,
\begin{equation}
y_f(x):=
\begin{cases}
0, & s_f(x)=0,\\
1, & s_f(x)=R,\\
\text{excluded}, & 0<s_f(x)<R.
\end{cases}
\label{eq:stable-endpoint-label}
\end{equation}
Thus the decision probes distinguish prompts that consistently
produce the correct call/no-call decision from prompts that
consistently exhibit the corresponding failure. Mixed-outcome
prompts are excluded from decision-basis construction, probe
fitting, and decision-probe evaluation. This setup evaluates
stable prompt-level failure propensity rather than stochastic
variation among repeated rollouts of one prompt.

\paragraph{Filtering, splitting, and class balance.}
For validity, an AST- and normalized-value-based judge compares
argument values with the reference on each parseable generated
call. For over-calling and missing, the stable label follows the
scenario type and whether a call was emitted across all rollouts.
Splits are grouped by prompt so that no prompt appears in multiple
splits. Contribution-basis construction uses equal-sized stable
failure and stable-correct development corpora, whereas detector
evaluation retains the natural class balance and reports ROC-AUC
and average precision. Under thinking, we split each response at
\texttt{</think>} and parse only the post-thinking answer, so a
tool mention inside reasoning is never counted as a call;
think-end features are taken strictly before \texttt{</think>} to
avoid leakage.

Table~\ref{tab:stable-endpoint-over} reports representative
over-calling prompt counts under the unified stable-endpoint
definition.
Retained is the sum of stable-correct and stable-failure prompts
after mixed outcomes are excluded, and $\pi_+$ is the
stable-failure prevalence among retained prompts.

\begin{table}[!t]
\centering
\scriptsize
\setlength{\tabcolsep}{3.6pt}
\caption{Representative over-calling stable-endpoint data audit.
Initial,
stable-correct (\textsc{sc}), mixed excluded, and stable-failure
(\textsc{sf}) are prompt counts.}
\label{tab:stable-endpoint-over}
\begin{tabular}{llrrrrrr}
\toprule
model & split & initial & \textsc{sc} & mixed & \textsc{sf} &
retained & $\pi_+$ \\
\midrule
Qwen3-4B & train & 829 & 554 & 103 & 172 & 726 & .237 \\
 & val & 178 & 110 & 18 & 50 & 160 & .313 \\
 & test & 177 & 119 & 22 & 36 & 155 & .232 \\
Qwen3-14B & train & 829 & 629 & 36 & 164 & 793 & .207 \\
 & val & 178 & 127 & 8 & 43 & 170 & .253 \\
 & test & 177 & 134 & 10 & 33 & 167 & .198 \\
Llama-3.1-8B & train & 1,878 & 1,563 & 109 & 206 & 1,769 & .116 \\
 & val & 402 & 337 & 26 & 39 & 376 & .104 \\
 & test & 404 & 324 & 27 & 53 & 377 & .141 \\
\bottomrule
\end{tabular}
\end{table}

The Qwen family produces sufficiently stable decision outcomes on
the canonical prompt pool, so this pool is adequate for fitting its
detectors; Qwen3-4B and Qwen3-14B are shown as representative
examples. For model families such as Llama and Gemma, some target
failures occur less frequently as stable endpoints, and we therefore
expand the construction pool to obtain enough examples; the Llama row
illustrates this case. This difference affects only detector
construction: for each model, the pre- and post-intervention results
are evaluated on the same fixed held-out evaluation set. It therefore
cannot explain the reported intervention effects.

Table~\ref{tab:basis-pool-audit} reports the size of each
contribution basis and the nonzero support retained by its
operational detector. As described above, each basis is estimated
from equal-sized stable-failure and stable-correct development
corpora.

\begin{table}[!t]
\centering
\footnotesize
\setlength{\tabcolsep}{4.2pt}
\caption{Contribution-basis and operational-detector audit.
Basis is the global $3\%$ union input dimension; support is the
exact nonzero $L_1$ support of the final operational detector.}
\label{tab:basis-pool-audit}
\begin{tabular}{llrr}
\toprule
model & signal & basis & support \\
\midrule
Qwen3-1.7B & over & 6,646 & 60 \\
 & miss & 7,097 & 10 \\
 & valid & 5,587 & 255 \\
Qwen3-4B & over & 13,440 & 51 \\
 & miss & 15,736 & 29 \\
 & valid & 11,126 & 330 \\
Qwen3-8B & over & 17,353 & 83 \\
 & miss & 20,285 & 110 \\
 & valid & 14,290 & 337 \\
Qwen3-14B & over & 27,963 & 51 \\
 & miss & 30,556 & 31 \\
 & valid & 22,553 & 742 \\
Llama-3.1-8B & over & 19,084 & 107 \\
 & miss & 16,587 & 10 \\
 & valid & 14,251 & 281 \\
Gemma-3-4B & over & 13,205 & 21 \\
 & miss & 12,262 & 16 \\
 & valid & 11,184 & 331 \\
\bottomrule
\end{tabular}
\end{table}

Validity is generated from $R=16$ source prompts and then filtered
to parseable calls. Table~\ref{tab:validity-source-audit} reports
the original rollout and parseable-call counts and the final calls
retained by the detector assets. BFCL is omitted from the table
because it is excluded from final pooled scoring.

\begin{table}[!t]
\centering
\scriptsize
\setlength{\tabcolsep}{3pt}
\caption{Source-wise validity filtering audit across all splits.
Invalid counts and prevalence are computed on the final retained
calls.}
\label{tab:validity-source-audit}
\begin{tabular}{llrrrrr}
\toprule
model & source & rollouts & parseable & retained & invalid & invalid rate \\
\midrule
Qwen3-1.7B & Glaive & 12,000 & 11,978 & 5,994 & 625 & .104 \\
 & ToolACE & 9,136 & 8,977 & 3,506 & 1,759 & .502 \\
 & xLAM & 12,000 & 11,896 & 5,965 & 2,224 & .373 \\
Qwen3-4B & Glaive & 12,000 & 11,968 & 5,996 & 611 & .102 \\
 & ToolACE & 9,136 & 8,437 & 3,430 & 1,689 & .492 \\
 & xLAM & 12,000 & 11,813 & 5,964 & 2,198 & .369 \\
Qwen3-8B & Glaive & 12,000 & 11,971 & 5,995 & 720 & .120 \\
 & ToolACE & 9,136 & 8,595 & 3,458 & 1,763 & .510 \\
 & xLAM & 12,000 & 11,890 & 5,968 & 2,124 & .356 \\
Qwen3-14B & Glaive & 12,000 & 11,761 & 5,968 & 692 & .116 \\
 & ToolACE & 9,136 & 8,126 & 3,353 & 1,726 & .515 \\
 & xLAM & 12,000 & 9,497 & 5,113 & 1,751 & .343 \\
Llama-3.1-8B & Glaive & 12,000 & 7,541 & 7,541 & 1,196 & .159 \\
 & ToolACE & 9,136 & 2,925 & 2,925 & 1,632 & .558 \\
 & xLAM & 12,000 & 6,146 & 6,146 & 2,274 & .370 \\
Gemma-3-4B & Glaive & 15,304 & 13,031 & 6,806 & 1,177 & .173 \\
 & ToolACE & 11,064 & 3,305 & 1,790 & 859 & .480 \\
 & xLAM & 15,096 & 11,801 & 6,241 & 2,272 & .364 \\
\bottomrule
\end{tabular}
\end{table}

\paragraph{Scope of validity and capability accuracy.}
Validity is evaluated only after a parseable call has been
generated. It does not decide whether a tool should be called, and
our experiments do not establish an automatic argument-repair or
regeneration procedure. Blocking, flagging, or regeneration are
possible downstream policies rather than evaluated outcomes.
Capability accuracy measures whether the model recovers a
parseable call to the required tool on prompts for which it
previously demonstrated the corresponding tool-use capability. A
call is counted as successful when it invokes the required tool
and contains the required argument fields. CA evaluates recovery
of tool invocation and selection; fine-grained argument
correctness is evaluated separately by the validity task.

\subsection{Prompt-Only and Reason-then-Act Controls}

We vary the system prompt style (guarded, neutral, or directive)
and whether thinking is enabled. Both are blunt global levers
rather than targeted interventions. Table~\ref{tab:promptbaseline}
shows that thinking induces an abstention bias for every prompt
style: over-calling nearly vanishes while missed calls rise.
Prompt wording only shifts the operating point and does not do so
monotonically. The emphatic directive gives the lowest
over-calling, whereas the terse guarded prompt gives the lowest
missing rate, because a strong ``MUST NOT call'' clause carries
over to tool-applicable cases. No wording escapes this trade-off,
motivating per-prompt detector gating. Argument validity is
unaffected by decision prompt style because it is evaluated on a
separate function-calling set without that system prompt.

\begin{table}[t]
\centering
\small
\caption{Prompt-only and Reason-then-Act controls (Qwen3-4B):
positive rate of each decision signal by prompt style and thinking
mode. Thinking trades over-calling for missed calls across all
styles; no style minimizes both.}
\label{tab:promptbaseline}
\begin{tabular}{lcccc}
\toprule
 & \multicolumn{2}{c}{over-calling} &
\multicolumn{2}{c}{missing} \\
\cmidrule(lr){2-3}\cmidrule(lr){4-5}
Prompt & no-think & thinking & no-think & thinking \\
\midrule
guarded   & 0.344 & 0.019 & \textbf{0.028} & 0.157 \\
neutral   & 0.297 & 0.007 & 0.173 & 0.191 \\
directive & \textbf{0.260} & \textbf{0.004} & 0.138 & 0.202 \\
\bottomrule
\end{tabular}
\end{table}

\subsection{Complete Reasoning-Mode Readout Results}
\label{app:reasoning-readouts}

Table~\ref{tab:think-full} reports every Qwen3 reasoning-mode
readout used to summarize migration in the main paper. Decision
features are compared at the prompt end and the reasoning end under
thinking; validity is read from the same generated call span under
no-think and thinking.

\begin{table}[h]
\centering
\small
\setlength{\tabcolsep}{4pt}
\caption{Complete chain-of-thought monitoring results.
\emph{Decisions} report per-prompt AUC/AP from prompt-end to
think-end readouts under thinking. \emph{Validity} reports
call-span AUC from no-think to thinking.}
\label{tab:think-full}
\begin{tabular}{@{}llccc@{}}
\toprule
Task & Model & earlier & later & $\Delta$ \\
\midrule
\multicolumn{5}{@{}l}{\emph{Decision readout: prompt-end $\to$ think-end (AUC / AP)}}\\
Over-calling & 1.7B & .975 / .907 & \textbf{1.00 / 1.00} & $+.09$ \\
             & 4B   & .609 / .083 & \textbf{.789 / .676} & $\mathbf{+.59}$ \\
             & 8B   & .927 / .415 & .743 / .674 & $+.26$ \\
             & 14B  & .987 / .667 & \textbf{1.00 / 1.00} & $+.33$ \\
Missing & 1.7B & .946 / .812 & \textbf{.997 / .991} & $+.18$ \\
        & 4B   & .970 / .831 & \textbf{.991 / .948} & $+.12$ \\
        & 8B   & .978 / .917 & \textbf{1.00 / 1.00} & $+.08$ \\
        & 14B  & .972 / .933 & \textbf{1.00 / 1.00} & $+.07$ \\
\midrule
\multicolumn{5}{@{}l}{\emph{Validity readout: no-think $\to$ thinking (AUC), same call span}}\\
Validity & 1.7B & .889 & \textbf{.913} & $+.024$ \\
         & 4B   & .886 & \textbf{.934} & $+.048$ \\
\bottomrule
\end{tabular}
\end{table}

\subsection{Intervention Conditions and Controls}

Let $N$ denote the number of layer--neuron pairs. We set
$k=m=\lceil0.03N\rceil$, retaining the top $3\%$ separately under
the failure and correct corpora before taking their union. Thus,
$3\%$ is an experimental setting rather than part of the method
definition. Monitoring uses held-out ROC-AUC and AP, with all
splits grouped by prompt.

For closed-loop intervention, we compare four operating
conditions: no steering, unconditional steering on every prompt,
detector-gated steering, and oracle-gated steering using the
ground-truth failure label. Oracle gating upper-bounds the benefit
of perfect failure detection. We report NTA, the erroneous-call
rate when no tool is supplied; DT, the erroneous-call rate when
only an inapplicable distractor is supplied; and capability
accuracy on tool-required prompts. The main gate figure summarizes
the first two with the prompt-count-weighted pooled rate
\[
\mathrm{OC}=
\frac{n_{\mathrm{NTA}}r_{\mathrm{NTA}}+
      n_{\mathrm{DT}}r_{\mathrm{DT}}}
     {n_{\mathrm{NTA}}+n_{\mathrm{DT}}},
\]
while the tables retain both strata. This distinguishes targeted
correction from a global shift in calling propensity.

For direction-level controls, we shuffle correct/failure labels,
reconstruct the direction, and repeat the intervention across
multiple seeds. Our matched-random selection control preserves
the selected-neuron count and per-layer profile, randomizes neuron
identity, constructs directions by the same procedure, and uses
the same relative-$\alpha$ schedule and token window. This is a
selection ablation rather than a strict statistical null because
randomly selected neurons can still encode the label-derived class
contrast. We also compare the contribution-first basis against
direct intervention on neurons selected only by probe
coefficients.

For the closed-loop gate, a \emph{shuffled-gate} control preserves
the intervention rate but permutes which prompts are steered. A
\emph{random-vector-in-real-gate} control retains the detector's
routing decisions but replaces each correcting direction with a
matched random vector. Together, these controls separate neuron
selection, steering direction, and prompt routing.

Finally, we apply a basic output-collapse screen to every
generation. It flags a response if its stripped output is empty,
or if it contains at least 12 whitespace-delimited tokens with a
unique-token ratio below $.35$. Intervention strengths that
trigger either empty or repetitively collapsed output are excluded
from claims about behavioral control. This screen tests for gross
generation failure rather than every possible change in output
quality.

\paragraph{What is shared and why.}
The term ``shared basis'' refers to sharing within a fixed failure
mode. For each $f$, the same contribution-selected set
$\mathcal{S}_f$ restricts the detector feature space and
constructs the layerwise intervention directions. The basis is not
one tied set shared across over-calling, missing, and validity:
each failure basis is constructed separately, although the
resulting sets may overlap. Contribution ranking first reduces the
all-layer feature space to
$\mathcal{S}_f$; the operational $L_1$ probe is then fit only
inside this reduced space. This avoids materializing and
optimizing each operational detector over the full set of
layer--neuron features.

\paragraph{Three distinct neuron counts.}
Three quantities serve different purposes and should not be
conflated. The smallest unconstrained sparse support measures the
readable signal's sparsity limit ($1$--$2$ neurons for missing,
$2$--$16$ for over-calling, and approximately $128$ for validity).
The operational detector uses a contribution-restricted input and
retains only its nonzero support at inference, typically tens of
features for decision failures and several hundred for validity.
Steering deliberately uses the broader
$3\%$ contribution basis, containing thousands of neurons, to
construct stable layerwise directions. The ``few-neuron'' result
therefore characterizes sparse monitoring; it does not claim that
every steering direction is constructed from only one or two
neurons.

\paragraph{Validity detector operating points.}
For Qwen3-4B validity, the full contribution basis contains 11,126
candidate features, and the fitted full-basis detector retains 330
nonzero coefficients while reaching $0.886$ AUC. The $K{=}4$ budget
contains 152 candidate features and reaches $0.841$ AUC. The
approximately 150-feature count and the $0.886$ AUC therefore
correspond to different operating points. Using the full-basis
operational support, the detector uses
$23\times$ fewer features than the 7,680-dimensional Healy baseline
and $279\times$ fewer features than the 92,160-dimensional
\textsc{When2Tool} baseline. Using operational supports consistently,
the reduction range across the main comparison table is
$23$--$627\times$.

\paragraph{Scope of the overlap diagnostic.}
The overlap analysis in the main paper was computed using the fixed
six-layer feature space retained from an earlier experimental
configuration. These layers define the reported common universe of
$6\times9{,}728=58{,}368$ layer--neuron pairs. We retain that table
as a restricted diagnostic and report a complete-space audit below.
The six-layer result also replicates qualitatively across solver
seeds: pairwise top-100 overlaps remain between zero and three, at
or below a layer-profile-matched null, and the three-way
intersection remains empty. Exact feature identities and exact-zero
counts should nevertheless not be interpreted as seed-invariant.
The main text therefore identifies this explicitly as a restricted
six-layer diagnostic rather than an all-layer independence result.

\paragraph{All-Layer Overlap Reveals a Small,
Configuration-Sensitive Shared Component.}
We extend the same $|L_1\text{ coefficient}|$ ranking to the
complete Qwen3-4B MLP feature space of
$36\times9{,}728=350{,}208$ layer--neuron features. Fits and
rankings use the training split only. The null independently
randomizes neuron identity within each layer while exactly
preserving each signal's selected count in that layer
($10{,}000$ draws). Full-space probe performance remains strong:
over-calling, missing, and validity reach AUC $.974$, $1.000$, and
$.869$, respectively, at the headline $C=.5$ setting.

\begin{table*}[t]
\centering
\scriptsize
\setlength{\tabcolsep}{3.8pt}
\caption{All-layer sparse-readout overlap for Qwen3-4B at
$C=.5$. Sets are ranked by absolute $L_1$ coefficient. The
hypergeometric expectation is $|A||B|/N$; the layer-matched null
preserves both sets' per-layer profiles. Sizes below $K$ reflect
the number of nonzero coefficients.}
\label{tab:overlap-all-layer}
\begin{tabular}{@{}rlrrrrrr@{}}
\toprule
$K$ & signal pair & sizes & obs. & hyp.\ exp. &
null mean & null 95\% interval & emp.\ $p$ \\
\midrule
100 & over-call $\cap$ missing  & 100/100 & 3 & .03 & .04 & [0,1] & .0002 \\
    & over-call $\cap$ validity & 100/100 & 2 & .03 & .03 & [0,1] & .0003 \\
    & missing $\cap$ validity   & 100/100 & 1 & .03 & .03 & [0,1] & .033 \\
\addlinespace
200 & over-call $\cap$ missing  & 200/200 & 3 & .11 & .16 & [0,1] & .0006 \\
    & over-call $\cap$ validity & 200/200 & 2 & .11 & .11 & [0,1] & .0069 \\
    & missing $\cap$ validity   & 200/200 & 1 & .11 & .12 & [0,1] & .110 \\
\addlinespace
500 & over-call $\cap$ missing  & 500/393 & 3 & .56 & .83 & [0,3] & .051 \\
    & over-call $\cap$ validity & 500/500 & 5 & .71 & .71 & [0,3] & .0013 \\
    & missing $\cap$ validity   & 393/500 & 3 & .56 & .55 & [0,2] & .018 \\
\addlinespace
1,000 & over-call $\cap$ missing  & 1,000/393 & 3 & 1.12 & 1.67 & [0,5] & .236 \\
      & over-call $\cap$ validity & 1,000/1,000 & 8 & 2.86 & 2.88 & [0,7] & .0097 \\
      & missing $\cap$ validity   & 393/1,000 & 3 & 1.12 & 1.16 & [0,4] & .110 \\
\midrule
\multicolumn{3}{@{}l}{Three-way intersection at $K=100/200/500/1{,}000$}
 & \multicolumn{5}{l@{}}{1 / 1 / 1 / 1
 \quad (empirical $p\leq.0045$)} \\
\bottomrule
\end{tabular}
\end{table*}

The complete-space audit preserves the magnitude-level separation
conclusion but refines the null-relative conclusion. Pairwise
Jaccard similarity never exceeds $.0153$, and no pair shares more
than eight selected features. Nevertheless, seven of the twelve
pairwise cells lie strictly above the layer-matched 95\% interval,
eight have an uncorrected one-sided empirical $p<.05$, and one
feature appears in all three sets at every budget. We therefore
describe these sparse readouts as largely failure-specific, not
literally disjoint. These permutation results are a descriptive
robustness audit rather than a familywise-corrected significance
test.

The shared component is configuration-sensitive. With stronger
regularization ($C=.05$), the test AUCs remain $.976$, $1.000$, and
$.881$, yet no pairwise cell exceeds its matched null and the
three-way intersection is empty at every budget. Changing only the
$C=.5$ solver seed produces similar small overlap counts but
different shared identities. Across those two fits, only $12\%$,
$36\%$, and $61\%$ of the over-calling, missing, and validity
top-100 sets reproduce, respectively. Thus the count-level pattern
is more reproducible than any particular shared neuron.

Where decision features overlap in the headline fit, over-calling
and missing read them with opposite polarity. This is consistent
with a general tool-invocation-versus-abstention signal whose
opposite deviations contribute to the two decision failures, while
the much larger nonoverlapping periphery may reflect
failure-specific information. The set overlap and coefficient
signs alone do not establish a unified geometric axis. The
all-layer depth profile supports the more limited conclusion that
decision information is late: only $16\%$ and $10\%$ of the
over-calling and missing top-100 features fall in the original
six-layer window, while $29\%$ and $70\%$ lie in layers 28--35.
For validity, the corresponding fractions are $15\%$ and $18\%$.

\begin{table}[t]
\centering
\scriptsize
\setlength{\tabcolsep}{3.2pt}
\caption{Ranking sensitivity at $K=100$. Entries in the overlap
column are over-call--missing / over-call--validity /
missing--validity. ``Above'' counts pairwise cells above the
layer-matched 95\% interval across all four $K$ budgets.}
\label{tab:overlap-ranking-sensitivity}
\begin{tabular}{@{}lrrr@{}}
\toprule
ranking & pairwise overlap & three-way & above \\
\midrule
$|L_1|$, $C=.5$       & 3/2/1   & 1  & 7/12 \\
$|L_1|$, $C=.05$      & 0/0/0   & 0  & 0/12 \\
$c^H$                 & 51/21/22 & 16 & 12/12 \\
$c^H-c^B$             & 3/8/7   & 2  & 12/12 \\
$|c^H-c^B|$           & 27/6/5  & 4  & 12/12 \\
layer-$z$ $c^H$       & 55/6/5  & 5  & 12/12 \\
\bottomrule
\end{tabular}
\end{table}

The ranking comparison distinguishes sparse failure readout from
the contribution rankings used to construct the broader
operational basis. Raw $c^H$, used to identify neurons with large
failure-corpus residual writes, has substantial cross-signal
overlap; within a signal it correlates $.946$ with its
correct-corpus counterpart $c^B$. It therefore primarily captures
a shared contribution-magnitude scaffold and should not itself be
read as a failure-specific coding score. Contrastive contribution
rankings also exceed the matched null, although much of their
support remains nonoverlapping. Restricted-space contribution
checks show the same ranking dependence, so it is not created by
expanding from six to 36 layers. These results do not alter the
role of each separately constructed failure basis in the
controller; they delimit what cross-failure circuit specificity can
be inferred from feature-set overlap alone.

\paragraph{Online gating and intervention implementation.}
Basis selection scores the unsigned magnitude of each neuron's
residual write, while detector features preserve the sign of the
SwiGLU activation at the input to \texttt{mlp.down\_proj}. At each
selected layer, the unit direction is scaled by
\[
\gamma^{(\ell)}
:=
\mathbb{E}_{(x,t)\in\mathcal{D}_{\mathrm{dev}}}
\left[
\left\|\operatorname{MLP}^{(\ell)}
(h_t^{(\ell)}(x))\right\|_2
\right],
\]
computed at the relevant readout positions on development data,
and then by the relative strength $\alpha$. The update is added to
the output of \texttt{mlp.down\_proj}, before the MLP branch joins
the residual stream. It is applied at the final prompt token and
the first five generated tokens at every layer with a nonempty
selected set.

The experimental controller first performs prompt scoring and
then a hooked generation pass. Because the decision readout
precedes generation, a deployment implementation may combine
these operations while reusing the prompt cache. Over-calling is
checked first; the missing branch is checked only if the
over-calling detector does not fire. Under thinking, the decision
score is read immediately before \texttt{</think>}.

\paragraph{Scope of the causal claim.}
Our interventions establish causal control at the level of
directions constructed from contribution-selected neuron writes:
adding these directions changes subsequent tool-call behavior,
while matched-random, label-shuffled, and probe-selected controls
do not reproduce the same effect. We do not claim that every
selected neuron is individually necessary, that the sparse
detector support alone is the model's native causal circuit, or
that decodability by itself establishes causal mediation.

\subsection{Calibration Sensitivity and Detector Conflicts}
\label{app:calibration-conflicts}

Detector fitting and basis construction use training prompts only.
We evaluate detector thresholds
$\tau\in\{.3,.5,.7,.9\}$ on prompt-disjoint validation and test
splits at $\alpha_{\rm down}=\alpha_{\rm up}=.2$. This common
setting was chosen using the validation-only operating-point
protocol described in the main paper and was then locked across all
six models. The evaluated suppression and elicitation grids are
$\{.05,.1,.2\}$ and $\{.1,.2\}$, respectively; the neighboring
grid points are reported only as sensitivity checks rather than as
model-specific operating points or test-set reselection. No
model-specific strength cap or generation safeguard was used.

The implementation checks suppression first if both detectors fire.
We audit the number and rate of such events on held-out prompts and
report whether this implementation choice affects any output.

\paragraph{Qwen3-4B threshold sensitivity.}
Table~\ref{tab:qwen4-threshold-sensitivity} reports the sensitivity
of the Qwen3-4B controller to the detector threshold. Performance is stable for
$\tau\in\{.3,.5,.7\}$. The locked $\tau=.7$ point retains the
highest observed test CA while intervening less often than
$\tau=.3$ or $.5$; at $\tau=.9$, uncorrected over-calling rises
substantially. Every routed condition has zero empty outputs and
zero repetition flags. The same serialized fitted detector objects
are used on validation and test. Suppression and elicitation IRs are the
mutually exclusive actions after suppression-priority routing;
\emph{both} counts raw detector co-firing before routing.

\begin{table*}[t]
\centering
\scriptsize
\setlength{\tabcolsep}{3.7pt}
\caption{Qwen3-4B failure-specific threshold sensitivity at locked
relative strength $.2/.2$. IR is intervention rate; OC is pooled
over-calling; CA is capability accuracy. The sweep is reported as a
sensitivity audit, not as test-set threshold selection.}
\label{tab:qwen4-threshold-sensitivity}
\begin{tabular}{@{}lrrrrrrrrrr@{}}
\toprule
split & $\tau$ & $N$ & supp.\ IR & elicit IR & total IR &
NTA & DT & OC & CA & both \\
\midrule
validation & .3 & 222 & .320 & .041 & .360 & .000 & .047 & .022 & .977 & 0 \\
validation & .5 & 222 & .297 & .036 & .333 & .000 & .058 & .028 & .977 & 0 \\
validation & .7 & 222 & .284 & .032 & .315 & .000 & .058 & .028 & .977 & 0 \\
validation & .9 & 222 & .230 & .032 & .261 & .076 & .058 & .067 & .977 & 0 \\
\midrule
test & .3 & 222 & .293 & .063 & .356 & .011 & .011 & .011 & .911 & 1 \\
test & .5 & 222 & .275 & .054 & .329 & .068 & .000 & .034 & .911 & 0 \\
test & .7 & 222 & .252 & .054 & .306 & .068 & .000 & .034 & .911 & 0 \\
test & .9 & 222 & .176 & .045 & .221 & .125 & .011 & .068 & .911 & 0 \\
\bottomrule
\end{tabular}
\end{table*}

\paragraph{Qwen3-4B dual-fire audit.}
At the locked $\tau=.7$ point, neither validation nor test contains
a dual fire. Consequently, suppression priority, missing priority,
and no-steering-on-conflict are empirically equivalent on these
streams; no additional conflict generation is required.

\begin{table}[t]
\centering
\small
\setlength{\tabcolsep}{7pt}
\caption{Qwen3-4B detector firing on the 222-prompt test
stream at $\tau_{\rm over}=\tau_{\rm miss}=.7$.}
\label{tab:qwen4-dual-fire}
\begin{tabular}{lrr}
\toprule
category & count & rate \\
\midrule
over-only & 56 & .252 \\
missing-only & 12 & .054 \\
neither & 154 & .694 \\
both-fire & 0 & .000 \\
\midrule
total & 222 & 1.000 \\
\bottomrule
\end{tabular}
\end{table}

\subsection{Complete Six-Model Sparsity Results}
\label{app:sparsity-results}

Table~\ref{tab:sparsity-full} expands the representative main-paper
table to all six evaluated models. Decision AUC is per prompt;
validity AUC is per rollout. Full-basis cells report AUC, the exact
selected-feature count and its percentage of the complete all-layer
MLP feature space, and the final detector's nonzero support. Other
cells report AUC and the actual union count.

\begin{table*}[t]
\centering
\footnotesize
\setlength{\tabcolsep}{2.5pt}
\caption{Complete detection AUC as the unified-basis budget shrinks.
$K$ is the per-layer-equivalent budget applied separately to the
error and correct rankings, so their union need not equal $K$ times
the number of layers. Full-basis cells contain three lines:
AUC; basis size (all-layer percentage); and nonzero support
(\emph{sup.}).}
\label{tab:sparsity-full}
\begin{tabular}{@{}llcccccc@{}}
\toprule
signal & budget & Qwen-1.7 & Qwen-4 & Qwen-8 & Qwen-14 & Llama-8 & Gemma-4 \\
\midrule
over- & full basis
 & \shortstack{.980\\6,646 (3.9\%)\\sup.\ 60} & \shortstack{.994\\13,440 (3.8\%)\\sup.\ 51}
 & \shortstack{.995\\17,353 (3.9\%)\\sup.\ 83} & \shortstack{.994\\27,963 (4.0\%)\\sup.\ 51}
 & \shortstack{.996\\19,084 (4.2\%)\\sup.\ 107} & \shortstack{.908\\13,205 (3.8\%)\\sup.\ 21} \\
call & $K{=}1$
 & .990 (35) & .996 (45) & .982 (44) & .984 (50)
 & .981 (52) & .966 (40) \\
\midrule
miss- & full basis
 & \shortstack{.996\\7,097 (4.1\%)\\sup.\ 10} & \shortstack{1.00\\15,736 (4.5\%)\\sup.\ 29}
 & \shortstack{1.00\\20,285 (4.6\%)\\sup.\ 110} & \shortstack{.997\\30,556 (4.4\%)\\sup.\ 31}
 & \shortstack{.895\\16,587 (3.6\%)\\sup.\ 10} & \shortstack{.927\\12,262 (3.5\%)\\sup.\ 16} \\
ing & $K{=}1$
 & .992 (38) & 1.00 (55) & 1.00 (54) & .990 (59)
 & .972 (43) & .811 (37) \\
\midrule
vali- & full basis
 & \shortstack{.889\\5,587 (3.2\%)\\sup.\ 255} & \shortstack{.886\\11,126 (3.2\%)\\sup.\ 330}
 & \shortstack{.895\\14,290 (3.2\%)\\sup.\ 337} & \shortstack{.856\\22,553 (3.2\%)\\sup.\ 742}
 & \shortstack{.856\\14,251 (3.1\%)\\sup.\ 281} & \shortstack{.868\\11,184 (3.2\%)\\sup.\ 331} \\
dity & $K{=}4$
 & .846 (119) & .841 (152) & .857 (156) & .833 (169)
 & .826 (136) & .816 (146) \\
 & $K{=}1$
 & .764 (28) & .780 (41) & .823 (40) & .825 (43)
 & .809 (35) & .765 (34) \\
\bottomrule
\end{tabular}
\end{table*}

\subsection{Complete Gated-Intervention Results}
\label{app:gate-results}

Table~\ref{tab:gate-full} decomposes the pooled over-call metric
used in the main figure. All six models use the same
validation-chosen and subsequently locked
$\alpha_{\rm down}=\alpha_{\rm up}=.2$ protocol setting. The
neighboring suppression and elicitation grid points are reported
only as supplementary sensitivity checks.

\paragraph{Qwen3-4B gate decomposition.}
Panel (c) of Table~\ref{tab:gate-full} compares all controller
components on the same $88/89/45$ NTA/DT/capability prompts.
Unconditional suppression and elicitation are separate because a
single prompt cannot receive both signs without a routing rule.
The oracle row routes baseline over-calling errors to the saved
unconditional-down output and baseline missing errors to the saved
unconditional-up output. Since generation is greedy and the hook
is row-independent, this cache composition is equivalent to an
oracle-routed pass at the same strength.

\begin{table*}[!t]
\centering
\scriptsize
\caption{Complete bidirectional gate results and controller
decomposition.
Panels (a--b) report all six models; NTA and DT are the
two over-calling strata; OC is their prompt-count-weighted pooled
rate. CA is capability accuracy. Shuffled preserves intervention
rate but changes prompt timing; random $d$ preserves the real gate
but replaces the intervention direction. Panel (c) decomposes the
Qwen3-4B controller at relative strength $.2$; IR is the fraction of
the 222 prompts edited, and collapse is the maximum rate under the
empty-output and repetitive-output checks.}
\label{tab:gate-full}
\begin{minipage}[t]{0.49\textwidth}
\centering
\setlength{\tabcolsep}{2.7pt}
\begin{tabular}{@{}llrrrr@{}}
\toprule
Model & Condition & NTA$\downarrow$ & DT$\downarrow$ &
OC$\downarrow$ & CA$\uparrow$ \\
\midrule
Qwen3-1.7B & baseline & .114 & .079 & .096 & .867 \\
 & real gate + real $d$ & \textbf{.000} & \textbf{.022} & \textbf{.011} & \textbf{.978} \\
 & shuffled gate & .136 & .112 & .124 & .800 \\
 & real gate + random $d$ & .136 & .079 & .107 & .889 \\
\midrule
Qwen3-4B & baseline & .534 & .011 & .271 & .756 \\
 & real gate + real $d$ & \textbf{.068} & \textbf{.000} & \textbf{.034} & \textbf{.911} \\
 & shuffled gate & .466 & .022 & .243 & .711 \\
 & real gate + random $d$ & .114 & .000 & .056 & .800 \\
\midrule
Qwen3-8B & baseline & .091 & .011 & .051 & .956 \\
 & real gate + real $d$ & \textbf{.000} & .011 & \textbf{.006} & \textbf{.978} \\
 & shuffled gate & .102 & .011 & .056 & .956 \\
 & real gate + random $d$ & .034 & .011 & .023 & .978 \\
\bottomrule
\end{tabular}

\vspace{4pt}
\textbf{(a) Qwen3-1.7B, 4B, and 8B}
\end{minipage}\hfill
\begin{minipage}[t]{0.49\textwidth}
\centering
\setlength{\tabcolsep}{2.7pt}
\begin{tabular}{@{}llrrrr@{}}
\toprule
Model & Condition & NTA$\downarrow$ & DT$\downarrow$ &
OC$\downarrow$ & CA$\uparrow$ \\
\midrule
Qwen3-14B & baseline & .420 & .011 & .215 & .791 \\
 & real gate + real $d$ & \textbf{.023} & .011 & \textbf{.017} & \textbf{.860} \\
 & shuffled gate & .352 & .022 & .186 & .767 \\
 & real gate + random $d$ & .284 & .011 & .147 & .814 \\
\midrule
Llama-3.1-8B & baseline & .010 & .228 & .117 & .685 \\
 & real gate + real $d$ & .010 & \textbf{.096} & \textbf{.052} & \textbf{.751} \\
 & shuffled gate & .016 & .228 & .119 & .671 \\
 & real gate + random $d$ & .010 & .234 & .119 & .637 \\
\midrule
Gemma-3-4B & baseline & .015 & .054 & .035 & .078 \\
 & real gate + real $d$ & .015 & .050 & .034 & \textbf{.508} \\
 & shuffled gate & .050 & .149 & .099 & .235 \\
 & real gate + random $d$ & .015 & .045 & .030 & .052 \\
\bottomrule
\end{tabular}

\vspace{4pt}
\textbf{(b) Qwen3-14B, Llama-3.1-8B, and Gemma-3-4B}
\end{minipage}

\vspace{7pt}
\begin{minipage}{\textwidth}
\centering
\setlength{\tabcolsep}{3.2pt}
\begin{tabular}{lrrrrrr}
\toprule
condition & NTA & DT & OC & CA & IR & collapse \\
\midrule
baseline & .534 & .011 & .271 & .756 & .000 & .000 \\
unconditional suppress & .011 & .000 & .006 & .511 & 1.000 & .000 \\
unconditional elicit & .773 & .382 & .576 & .911 & 1.000 & .000 \\
detector-gated & .068 & .000 & .034 & .911 & .279 & .000 \\
oracle-gated & .011 & .000 & .006 & .911 & .261 & .000 \\
\bottomrule
\end{tabular}

\vspace{4pt}
\textbf{(c) Qwen3-4B controller decomposition}
\end{minipage}
\end{table*}

\section{Additional Reproducibility Details}
\label{app:reproducibility-details}

The code and processed natural-rollout datasets used in this work
will be released publicly upon publication under a license permitting
research use. This section records the settings needed to
reconstruct the sparse detectors and intervention pipeline.

\subsection{Hyperparameter Search Space and Selection}

\paragraph{Generation and rollout parameters.}
Decision datasets use $R=8$ stochastic rollouts per prompt and
validity uses $R=16$, with temperature $0.7$ and each checkpoint's
native chat template. Qwen3 thinking-mode experiments use the
model's native thinking configuration; all other primary results
use no-think generation. Rollout generation uses top-$p=.95$ and
at most $512$ new tokens. Intervention evaluation is greedy and
uses at most $128$ new tokens. The checkpoint identifiers are
\texttt{Qwen/Qwen3-\{1.7B,4B,8B,14B\}},
\texttt{meta-llama/Llama-3.1-8B-Instruct}, and
\texttt{google/gemma-3-4b-it}.

\paragraph{Contribution-basis construction.}
For $N$ layer--neuron pairs, we use
$k=m=\lceil.03N\rceil$ and take the union of the globally
top-ranked failure and stable-correct contributions. The ranking
uses
$|a_i^{(\ell)}|\|v_i^{(\ell)}\|_2$, where
$a_i^{(\ell)}$ is the signed SwiGLU gated product at the input to
\texttt{mlp.down\_proj} and
$v_i^{(\ell)}$ is the corresponding down-projection column.
There is no per-layer quota. Restricted top-$K$ probes refit their
regularization on validation data; the reported sufficient size is
the smallest $K$ satisfying
$\mathrm{AUC}_K\geq.95\,\mathrm{AUC}_{\mathrm{full}}$.
The complete sweep is
$K\in\{1,2,4,8,16,32,64,128,256,512\}$, in addition to the full
$3\%$ basis. The locked $3\%$ setting is used in the final
cross-model tables. Signed CETT
divides by
$\max(\|\mathrm{MLP}^{(\ell)}_{\rm out}\|_2,10^{-6})$.
Direction normalization separately uses a $10^{-8}$ norm floor.

\paragraph{Probe fitting.}
The operational detector is an $L_1$-regularized logistic
regression fit on signed, layer-normalized features inside the
selected basis. Hyperparameters are selected using training and
validation data only, and the held-out test set is evaluated once
after freezing the configuration.
Each feature is standardized with a \texttt{StandardScaler} fit on
the training split. The classifier uses the \texttt{liblinear}
solver, balanced class weights, and
$C\in\{.01,.03,.1,.3\}$. Candidate values are selected using
validation ROC--AUC and thresholded F1. Reported support counts the
nonzero fitted coefficients.

The matched dense residual baseline uses standardized features and
an $L_2$ logistic regression with $C\in\{.01,.1,1\}$. The
all-layer \textsc{When2Tool} reproduction uses the published
classifier configuration.

\paragraph{Steering and gate calibration.}
Directions are injected at the output of
\texttt{mlp.down\_proj} for the final prompt token and first five
generated tokens. The common closed-loop operating point is
$\tau_{\rm over}=\tau_{\rm miss}=.7$ and
$\alpha_{\rm over}=\alpha_{\rm miss}=.2$ in the submitted protocol.
The threshold audit uses
$\{.3,.5,.7,.9\}$. Across the six-model closed-loop sweeps, the
suppression grid is
$\alpha_{\rm down}\in\{.05,.1,.2\}$ and the elicitation grid is
$\alpha_{\rm up}\in\{.1,.2\}$. The common
$\alpha_{\rm down}=\alpha_{\rm up}=.2$ setting was chosen using the
validation-only operating-point protocol described in the main
paper and was then locked across all six models. The neighboring
grid points are reported only as sensitivity checks rather than as
model-specific operating points or test-set reselection.
Over-calling is checked first if both detectors fire.
The basic output-collapse screen flags stripped empty responses
and responses of at least 12 whitespace-delimited tokens whose
unique-token ratio is below $.35$.
No retry, fallback, or additional model-specific generation
safeguard was found in the audited implementation.

\subsection{Randomness and Repeated Controls}

Train, validation, and test splits are grouped by prompt and are
fixed before model fitting. Model rollouts are stochastic;
contribution ranking and deterministic probe fitting introduce no
additional randomness once their inputs and solver state are
fixed. Randomized controls independently resample the relevant
component while preserving the comparison's remaining structure:
matched-random neuron selection preserves the selected count and
per-layer profile, random directions are renormalized and use the
same layer scales, label shuffling reconstructs the direction
after permuting labels, and gate shuffling preserves the
intervention count.

\paragraph{Repeated randomized controls.}
Prompt splits are fixed before fitting. Primary detector and
steering results use one frozen split and one evaluation pass.
Matched-random detection uses five draws in the main tables and ten
draws in the representative Qwen3-4B audit. The representative
matched-random steering audit uses five draws. Label-shuffle audits
use 20 permutations, except for the explicitly stated three-run
resource-limited cases. Shuffled-gate and random-vector rows are
single control draws rather than stability intervals.

\paragraph{Representative resampling audit.}
To keep the robustness analysis computationally focused, we use
Qwen3-4B as the representative model for additional random-basis
and matched-random-direction resampling. The six-model tables
retain the originally reported fixed-split results.
Table~\ref{tab:qwen4-random-basis-existing} reports the five draws
originally saved together with five additional independently
resampled draws. The table reports population standard deviations
and ranges across all ten draws for both
AUC and AP. Table~\ref{tab:qwen4-random-steering} separately
reports five matched-random steering draws.

\begin{table*}[t]
\centering
\scriptsize
\setlength{\tabcolsep}{3.2pt}
\caption{Qwen3-4B layer-profile-matched random-subset stability
over ten draws. Standard deviations are population standard
deviations; brackets give the draw-wise minimum and maximum.}
\label{tab:qwen4-random-basis-existing}
\begin{tabular}{llrrrrr}
\toprule
signal & budget & features & AUC mean$\pm$sd & AUC range &
AP mean$\pm$sd & AP range \\
\midrule
over-calling & full & 13,440 & $.9938\pm.0009$ & $.9926$--$.9949$ & $.9876\pm.0019$ & $.9848$--$.9902$ \\
 & $K=16$ & 751 & $.9930\pm.0011$ & $.9912$--$.9948$ & $.9855\pm.0028$ & $.9813$--$.9897$ \\
 & $K=4$  & 184 & $.9918\pm.0020$ & $.9894$--$.9942$ & $.9827\pm.0046$ & $.9772$--$.9885$ \\
 & $K=1$  & 45  & $.9893\pm.0050$ & $.9797$--$.9970$ & $.9755\pm.0149$ & $.9425$--$.9941$ \\
missing & full & 15,736 & $1.0000\pm.0000$ & $1.0000$--$1.0000$ & $1.0000\pm.0000$ & $1.0000$--$1.0000$ \\
 & $K=16$ & 920 & $.9992\pm.0012$ & $.9973$--$1.0000$ & $.9977\pm.0035$ & $.9924$--$1.0000$ \\
 & $K=4$  & 232 & $1.0000\pm.0000$ & $1.0000$--$1.0000$ & $1.0000\pm.0000$ & $1.0000$--$1.0000$ \\
 & $K=1$  & 55  & $.9989\pm.0018$ & $.9947$--$1.0000$ & $.9971\pm.0047$ & $.9860$--$1.0000$ \\
validity & full & 11,126 & $.8739\pm.0052$ & $.8626$--$.8822$ & $.7396\pm.0127$ & $.7240$--$.7553$ \\
 & $K=16$ & 611 & $.8551\pm.0088$ & $.8425$--$.8720$ & $.6870\pm.0138$ & $.6684$--$.7133$ \\
 & $K=4$  & 152 & $.8008\pm.0169$ & $.7704$--$.8254$ & $.6050\pm.0367$ & $.5472$--$.6478$ \\
 & $K=1$  & 41  & $.7476\pm.0287$ & $.7032$--$.7925$ & $.5282\pm.0402$ & $.4729$--$.6115$ \\
\bottomrule
\end{tabular}
\end{table*}

\begin{table*}[t]
\centering
\scriptsize
\setlength{\tabcolsep}{4.2pt}
\caption{Mean $\pm$ population standard deviation for Qwen3-4B
unconditional matched-random-basis suppression over five draws at
relative $\alpha=.2$. Every run edits all 222 prompts using 13,440
neurons while preserving the real basis's layer profile. Empty and
Rep.\ are the implemented empty-output and repetition proxies;
Malf.\ is malformed-call rate.}
\label{tab:qwen4-random-steering}
\begin{tabular}{lrrrrrrr}
\toprule
condition & NTA$\downarrow$ & DT$\downarrow$ & OC$\downarrow$ &
CA$\uparrow$ & Empty & Rep. & Malf. \\
\midrule
matched random &
$.3227\pm.2319$ & $.0135\pm.0218$ & $.1672\pm.1221$ &
$.6800\pm.0896$ & $.0000\pm.0000$ & $.0000\pm.0000$ &
$.0162\pm.0199$ \\
\bottomrule
\end{tabular}
\end{table*}

The random-basis detection audit is stable at the full budget:
over-calling remains at $.9938\pm.0009$ AUC, missing is perfect,
and validity reaches $.8739\pm.0052$. Validity is more sensitive
at the one-neuron-per-layer-equivalent budget
($.7476\pm.0287$). Matched-random steering, however, varies
substantially across draws (OC standard deviation $.1221$ and CA
standard deviation $.0896$). This control preserves the real
basis's per-layer allocation, so it retains coarse information
about which layers are emphasized, and randomly sampled neurons in
those layers can still carry part of the relevant signal.
Nevertheless, its mean residual OC is $.1672$, compared with
$.006$ for the contribution-selected unconditional suppress
direction. The detector-gated contribution-selected controller
further attains OC $.034$ with CA $.911$. Thus, matched-random
directions can be behaviorally active but are substantially less
effective and much less reliable than the systematically
constructed direction. Empty-output and repetition rates remain
zero, while malformed calls average $.0162\pm.0199$.

\subsection{Software Environment}

The recorded environment uses Python 3.12.13, PyTorch 2.8.0 with
CUDA 12.8, Transformers 4.56.1,
scikit-learn 1.8.0, NumPy 2.2.6, Accelerate 1.13.0, and vLLM
0.11.0. Model forward passes use bfloat16.

\subsection{Runs, Uncertainty, and Statistical Tests}

Detection AUC and AP use prompt as the evaluation unit for decision
failures and retained generated calls for validity. The primary
six-model tables report the frozen train/validation/test split
rather than treating repeated rollouts of one prompt as independent
decision examples. Additional random-basis variation is reported
only for the representative Qwen3-4B audit described above.

For the paired \textsc{When2Tool} suppression analysis, we use
two-sided exact McNemar tests on examples whose baseline call is
suppressed; the corresponding sample sizes and $p$ values appear
in Table~\ref{tab:w2t-paired-tradeoff}. Random-direction,
label-shuffle, and gate-shuffle experiments are control
interventions rather than substitutes for uncertainty over the
full model population.

\paragraph{Run and uncertainty inventory.}
Each real detection cell is one selected fit on one fixed split.
Each real direct-steering or gated-steering condition is one greedy
generation pass over the fixed prompt stream. Repeated randomized controls report the number
of draws stated above. No prompt-bootstrap confidence intervals or
multiple-comparison correction are reported for the primary
six-model tables.

\subsection{Computational Cost and Detector-Side Overhead}

Contribution selection reduces the feature space before fitting
the operational detector, and the fitted linear heads retain sparse
nonzero supports. The current prototype gathers the selected
contribution bases before applying these sparse heads, so the timing
reported below includes this feature-gather cost. These reductions
do not imply an equal-factor reduction in the cost of the underlying
language-model forward pass.
For Qwen3-4B, the all-layer space contains $350{,}208$ MLP
features. The over-calling, missing, and validity bases contain
$13{,}440$, $15{,}736$, and $11{,}126$ features, respectively;
their final operational supports contain $51$, $29$, and $330$ nonzero
coefficients. These counts describe detector-side reduction rather
than end-to-end language-model acceleration.

In a representative Qwen3-4B component-level benchmark, the sparse
detector head requires approximately $1.6$\,ms per prompt, or
$0.25\%$ of the corresponding unsteered generation time. Applying
the gated steering vectors through the current prototype hooks adds
approximately $141$\,ms per prompt (a median overhead of $21.6\%$)
relative to unsteered generation, without a measurable increase in
peak allocated memory. The latter figure includes overhead from the
current unfused hook implementation and should not be interpreted as
the cost of the vector addition alone. These component-level timings
exclude the language-model forward pass used to obtain detector
features and therefore do not represent end-to-end controller
latency.

\paragraph{AI Use Declaration.}
Generative AI tools were used for minor language polishing, manuscript
organization, and \LaTeX{} preparation. The authors verified all
technical content, experimental results, and references.

\FloatBarrier
\section{Detailed Comparison and Transfer on \textsc{When2Tool}}
\label{app:baseline-comparison}

\subsection{Matched Detector Comparison}
\label{app:w2t-detector}

\textsc{When2Tool} reads last-input-token residual states with an
$L_2$-regularized linear probe \citep{sun2026when2tool}. We first
compare detectors using its original prompt wrapper, the same
$900/2{,}250$ train/test split, and identical necessity labels and
scorer. Our sparse CETT probe reaches $0.934$ pooled AUROC with
$151$ nonzero inputs, versus $0.880$ for the $94{,}720$-dimensional
dense probe, and leads at every difficulty
(Table~\ref{tab:necessity_diff}). The dense probe nearly saturates
training AUROC ($0.9998$), whereas the sparse probe generalizes
better from the same $900$ training prompts.

\begin{table}[t]
\centering
\small
\setlength{\tabcolsep}{4pt}
\caption{Matched necessity detection on \textsc{When2Tool}'s
original prompt wrapper (Qwen3-4B). Both probes use the same
$900/2{,}250$ split, labels, and scorer.}
\label{tab:necessity_diff}
\begin{tabular}{lccc}
\toprule
difficulty & sparse CETT & dense residual & $\Delta$ \\
\midrule
easy    & \textbf{0.916} & 0.874 & $+0.042$ \\
medium  & \textbf{0.871} & 0.812 & $+0.059$ \\
hard    & \textbf{0.933} & 0.847 & $+0.086$ \\
\midrule
pooled  & \textbf{0.934} & 0.880 & $+0.055$ \\
\bottomrule
\end{tabular}
\end{table}

The ordering is robust to prompt format. Under our wrapper, the
sparse and dense probes reach $0.937$ and $0.868$, respectively,
with gains at every difficulty. Thus the difference is not
explained by either labels or prompt formatting. This comparison
is limited to the faithfully reproduced benchmark and does not
imply that dense residual probes are uniformly weak.

\subsection{Replacing the Gate while Retaining Textual Prefill}
\label{app:w2t-gate-swap}

To isolate detector quality from the intervention mechanism, we
retain \textsc{When2Tool}'s textual-prefill actuator and replace
only its dense probability estimate with our sparse detector.
At matched average tool-call rates, the sparse gate raises robust
accuracy by $1.7$--$2.2$ points throughout the evaluated frontier
(Table~\ref{tab:w2t-gate-swap}). This controlled swap shows that
detection quality is a practical bottleneck even when the
original actuator is left unchanged.

\begin{table}[t]
\centering
\small
\setlength{\tabcolsep}{4pt}
\caption{Gate substitution with the textual-prefill actuator
fixed. Cells report robust accuracy at matched average tool calls
on the $750$-prompt stratified subset.}
\label{tab:w2t-gate-swap}
\begin{tabular}{ccc}
\toprule
avg.\ tool calls & dense gate & sparse gate \\
\midrule
.587 & .727 & \textbf{.744} \\
.516 & .706 & \textbf{.728} \\
.457 & .677 & \textbf{.699} \\
\bottomrule
\end{tabular}
\end{table}

\subsection{Cross-Corpus Transfer of the Steering Direction}
\label{app:w2t-transfer}

We next test whether the actuator itself transfers. A suppress
direction estimated on our natural over-calling corpus is applied
to \textsc{When2Tool} without refitting. The source corpus defines
failure as calling a nonexistent tool, whereas \textsc{When2Tool}
labels benchmark-level unnecessary calls. Despite this mismatch,
the transferred direction lowers the misuse rate by $18.5$ points
relative to its layer-profile-matched random direction at
$\alpha=.4$, with no observed degeneration
(Table~\ref{tab:w2t-transfer}). A direction estimated locally on
\textsc{When2Tool} is also specific but weaker at the same dose.

\begin{table}[t]
\centering
\scriptsize
\setlength{\tabcolsep}{2.2pt}
\caption{Suppress-direction transfer at $\alpha=.4$. ``Random''
uses a matched random direction; gap is real minus random.
Accuracy uses the robust scorer.}
\label{tab:w2t-transfer}
\begin{tabular}{lrrrrrr}
\toprule
direction source & misuse & random & gap & acc. & cover & degen. \\
\midrule
\textsc{When2Tool} local & .694 & .793 & $-.099$ & .799 & .859 & .000 \\
natural over-calling & \textbf{.617} & .802 & $\mathbf{-.185}$ & .769 & .850 & .000 \\
\bottomrule
\end{tabular}
\end{table}

At comparable accuracy cost, the local and transferred directions
reach similar points: local $\alpha=.5$ gives a $19.7$-point net
reduction for a $4.1$-point accuracy decrease, while transferred
$\alpha=.4$ gives an $18.5$-point reduction for a $3.8$-point
decrease. We therefore interpret this result as cross-corpus
transfer of a tool-calling tendency, not evidence for a universal
tool-use direction.

\subsection{Intervention Frontier and Causal Controls}
\label{app:w2t-frontier}

Table~\ref{tab:w2t-alpha-frontier} reports the complete
effect--cost frontier for the local per-layer direction. The real
direction separates increasingly from the matched random control
through $\alpha=.6$ without degeneration; stronger intervention
reveals a degeneration cliff. For $\alpha\leq.4$, accuracy on
tool-required prompts remains exactly at the baseline value
($0.808$), so the low-dose effect does not spill into the need
side.

\begin{table}[t]
\centering
\scriptsize
\setlength{\tabcolsep}{2.2pt}
\caption{Per-layer suppress-direction frontier. Net is real minus
matched random misuse. Degeneration is measured among touched
prompts.}
\label{tab:w2t-alpha-frontier}
\begin{tabular}{crrrrrr}
\toprule
$\alpha$ & misuse & random & net & acc. & cover & degen. \\
\midrule
.0 & .802 & ---  & ---    & .807 & .866 & .000 \\
.1 & .750 & .799 & $-.049$ & .804 & .862 & .000 \\
.2 & .728 & .799 & $-.071$ & .812 & .859 & .000 \\
.4 & .694 & .793 & $-.099$ & .799 & .857 & .000 \\
.5 & .599 & .796 & $-.197$ & .765 & .854 & .000 \\
.6 & .429 & .796 & $-.367$ & .729 & .854 & .000 \\
.7 & .358 & .812 & $-.454$ & .691 & .852 & .024 \\
\bottomrule
\end{tabular}
\end{table}

The routing controls separate \emph{when} from \emph{how}
(Table~\ref{tab:w2t-controls}). A random direction under the real
gate is inert, whereas shuffling the gate weakens suppression,
damages need-side coverage, and introduces degeneration. Thus the
direction supplies behavioral leverage and the detector supplies
selectivity.

\begin{table}[t]
\centering
\small
\setlength{\tabcolsep}{3pt}
\caption{Causal controls at $\alpha=.6$ (per-layer direction).}
\label{tab:w2t-controls}
\begin{tabular}{lrrrr}
\toprule
condition & misuse & cover & acc. & degen. \\
\midrule
real gate + real direction & \textbf{.429} & .854 & .729 & .000 \\
shuffled gate              & .627 & .721 & .695 & .062 \\
real gate + random direction & .796 & .864 & \textbf{.800} & .000 \\
\bottomrule
\end{tabular}
\end{table}

Per-layer selection is also materially safer than global
selection on this benchmark. The global rule concentrates
$55.5\%$ of its budget in the final five layers, becomes
indistinguishable from its random control at high dose, and
degenerates one step earlier. The per-layer rule keeps its random
control flat and preserves a specific effect. The concentration
pattern is a plausible explanation, but we do not claim it as a
separately identified mechanism.

\subsection{Benchmark-Dependent Accuracy Trade-Off}
\label{app:w2t-benchmark-tradeoff}

\textsc{When2Tool}'s necessity labels are task-level and
model-independent, whereas whether a tool helps can depend on
model capability. On paired prompts that the baseline calls but
an intervention suppresses, answering without the tool
significantly reduces robust accuracy under both steering and
prefill (Table~\ref{tab:w2t-paired-tradeoff}). This explains why
all call-suppression methods face a nontrivial accuracy cost in
this Qwen3-4B evaluation.

\begin{table}[t]
\centering
\scriptsize
\setlength{\tabcolsep}{3pt}
\caption{Paired accuracy on examples labeled unnecessary whose
baseline call is suppressed. Exact McNemar tests are two-sided.}
\label{tab:w2t-paired-tradeoff}
\begin{tabular}{lrrrrr}
\toprule
intervention & $n$ & called & suppressed & $\Delta$ & $p$ \\
\midrule
steering ($\alpha=.4$) & 35  & .829 & .457 & $-.371$ & $9.8{\times}10^{-4}$ \\
prefill ($\tau=.5$)    & 218 & .885 & .693 & $-.193$ & $5.7{\times}10^{-8}$ \\
\bottomrule
\end{tabular}
\end{table}

\subsection{Scoring Protocol and Scope}
\label{app:w2t-scoring}

All intervention accuracies above use a robust answer extractor:
after removing any control prefix, it checks boxed answers,
``answer is'' constructions, the final nonempty line, and a final
literal through \textsc{When2Tool}'s value comparator. The
original exact whole-response fallback penalizes explanatory
answers induced by textual prefill and also changes baseline
scores. Manual inspection of all $68$ baseline corrections found
them to be genuine formatting-equivalent answers. We therefore
exclude strict-extraction accuracy from all claims.

We also determine the steering sign from the semantic roles of
the source corpora: the correct-minus-failure direction suppresses
calling when the correct side abstains. All reported intervention
results use this corrected convention; exploratory runs with the
opposite sign are excluded. The transfer and frontier analyses are
currently limited to Qwen3-4B and one external benchmark, and a
two-tailed prefill prototype is excluded because it does not match
the suppress-only touched set.

\subsection{Comparison with a Prior Validity Detector}
\label{app:validity-detector}

For the internal-representation validity detector of
\citet{healy2026internal}, per-source comparisons are more
informative than the pooled score because source prevalence is
heterogeneous. We outperform the reproduced recipe on Glaive
($0.905$ vs.\ $0.869$) but trail it on xLAM ($0.817$ vs.\
$0.863$) and ToolACE ($0.823$ vs.\ $0.846$). Running their
detector unchanged on our natural rollout-error labels raises its
AUROC from their reported $0.721$ on mask-and-replace labels to
$0.856$. This cross-paper gap also includes a model change; the
controlled claim is therefore the per-source comparison above.

\FloatBarrier
\section{Extended Related Work}
\label{app:related-work}

\paragraph{Tool use and function calling in LLMs.}
Function-calling research has largely focused on improving and behaviorally
evaluating whether models select the right API and generate valid arguments.
BFCL introduced AST- and execution-based evaluation across simple, parallel,
and relevance-detection settings \citep{patil2025bfcl}; ToolACE and xLAM
developed scalable data pipelines and specialized models for these tasks
\citep{liu2025toolace,zhang2025xlam}, while Glaive released a public
function-calling corpus \citep{glaive2023function}. Reliability-oriented work
further distinguishes tool-selection from tool-usage hallucinations and trains
models to defer or seek clarification \citep{xu2025reliability}. Most recently,
\citet{yin2026reasoningtrap} show that stronger explicit reasoning can amplify
tool hallucination, while KATE improves multi-step execution by integrating
experiential knowledge across inference and post-training
\citep{hao2026kate}. These studies measure or train observable tool behavior;
we instead ask where naturally occurring validity, over-calling, and missing
failures are represented and whether those representations are causally
controllable.

\paragraph{Probing internal representations.}
Linear probes have shown that hidden states encode information about
truthfulness even when model outputs are wrong \citep{azaria2023lying}.
This idea has recently reached tool-using agents. Dense residual-stream
features can detect incorrect tool selection and malformed parameters
\citep{healy2026internal}, and tool necessity is decodable before generation
\citep{sun2026when2tool}. Concurrent work also finds that tool identity is
linearly readable and steerable across model families \citep{wu2026toolcalling}.
A complementary model-adaptive analysis separates a model's internal judgment
of tool necessity from its executed action and finds that the two probe
directions become nearly orthogonal at the late-layer readout
\citep{cheng2026knowingdoing}.
We extend this line from dense, task-level representations to signed activations
of individual MLP neurons, distinguish three failure modes at their appropriate
readout locations, and quantify both layer locality and the minimum sufficient
support.

\paragraph{Sparse neurons and contribution-based localization.}
Our selection score is related to Wanda's activation-weight criterion for
identifying consequential parameters \citep{sun2024wanda}, but we rank raw MLP
neurons globally across layers to construct a behavioral intervention basis.
This grouped use of WANDA also connects to safety attribution, where it has
identified sparse safety-critical regions whose removal breaks refusal behavior
\citep{wei2024brittleness}.
Sparse neuron-level mechanisms have been reported for factual hallucination:
H-Neurons use fewer than $0.1\%$ of neurons to predict hallucinations and link
them causally to over-compliance \citep{gao2025hneuron}. In safety, manipulating
knowledge neurons controls refusal and motivates neuron-specific SafeTuning
\citep{zhao2026safety}, while single-neuron interventions can bypass refusal
alignment \citep{kazemi2026single}. These results establish fine-grained
localization for factuality and safety; our focus is tool-use reliability and a
single contribution-selected basis shared by monitoring and intervention.

\paragraph{Activation steering and causal intervention.}
Representation engineering and activation addition control high-level behavior
by shifting internal states at inference time
\citep{zou2023representation,turner2023activation}; ITI learns sparse
attention-head directions that improve truthfulness \citep{li2023iti}.
Agent-specific extensions steer residual states to reduce overthinking and
overacting \citep{sui2026tact}, while ASA uses router-conditioned, probe-gated
activation steering for training-free tool-domain adaptation
\citep{wang2026asa}. Very recent work demonstrates bidirectional control of
tool invocation from heading-position steering vectors, while also finding
that tool-use geometry is diffuse and tool-dependent
\citep{chen2026heading}. In contrast to unconditional or task-wide steering,
\textsc{PRISMS} constructs an independent direction at each selected layer from
the same neuron basis used for detection, separates over-calling from missing,
and gates each direction with its corresponding sparse failure detector.

\end{document}